\documentclass[letterpaper, 10 pt, conference]{ieeeconf}  

\IEEEoverridecommandlockouts                              

\usepackage{times}
\usepackage{amsfonts}
\usepackage{color}
\usepackage{graphicx}
\usepackage{amsmath}
\usepackage{amssymb}
\usepackage[table]{xcolor}


\usepackage[ruled, vlined,linesnumbered]{algorithm2e}
\usepackage[caption=false]{subfig}

\title{\LARGE \bf
Analyzing the Utility of a Support Pin in \\
Sequential Robotic Manipulation
}

\author{Chao Cao, Weiwei Wan, Jia Pan, and Kensuke Harada
\thanks{Chao Cao is with the Department of Computer Science, the University of Hong Kong.
        Jia Pan is with the Department of Mechanical and Biomedical Engineering, the City University of Hong Kong.
Weiwei Wan and Kensuke Harada are with National Institute of Advanced Industrial Science and Technology (AIST), Japan.
        }%
}

\begin{document}

\maketitle
\thispagestyle{empty}
\pagestyle{empty}

\begin{abstract}
Pick-and-place regrasp is an important manipulation skill for a
robot. It helps a robot accomplish tasks that cannot be achieved within a single
grasp, due to constraints such as kinematics or collisions between the
robot and the environment. Previous work on pick-and-place regrasp only
leveraged flat surfaces for intermediate placements, and thus is limited in the
capability to reorient an object.

In this paper, we extend the reorientation capability of a pick-and-place
regrasp by adding a vertical pin on the working surface and using it as the
intermediate location for regrasping. In particular, our method automatically
computes the stable placements of an object leaning against a vertical pin,
finds several force-closure grasps, generates a graph of regrasp actions, and
searches for the regrasp sequence. To compare the regrasping performance with and
without using pins, we evaluate the success rate and the length of regrasp
sequences while performing tasks on various models. Experiments on reorientation
and assembly tasks validate the benefit of using support pins for regrasping.
\end{abstract}

\section{Introduction}
\label{sec:intro}

To rearrange and interact with a scene, a robot needs to be able to grasp and
manipulate objects. One of the most common manipulation tasks is the
pick-and-place, where the robot picks up a target object at an initial pose and
then places it at a target pose. Sometimes the desired target pose is not
reachable directly, due to the robot's kinematics constraints or collisions
between the robot and its surrounding environment. A typical solution is the pick-and-place regrasp, i.e., the robot uses a sequence of
pick-ups and place-downs to incrementally change the object's pose. In
particular, after the object is picked up by the first grasp, it is stably
placed in an intermediate location and then picked up again using another grasp.
The relative pose between the object and the manipulator is fixed during each
grasp, and only changes when the robot places the object down and regrasps
it~\cite{Tournassoud:1987:R}. It is desirable if the object has many different
ways of placements and each placement has many valid grasps, because this can
provide the robot with more choices to incrementally adjust the object's pose. 
More formally, the flexibility in placements helps to increase the
connectivity of the regrasp graph~\cite{Wan:ICRA:2015} and is crucial for the
quality of the resulting regrasp sequence.

There has been extensive work~\cite{Lozano-Perez:1992:HRT,Terasaki:1998:MPI} on
pick-and-place regrasp since the 1980s, due to its importance for single arm
manipulation. The majority of previous work assumed flat intermediate placement location,
e.g., a horizontal ground or a tilted table, and focused on computing a feasible
or optimal trajectory in the high-dimensional configuration space for achieving
robotic pick-and-place tasks. However, since the convex hull of most objects
only has limited number of faces, these objects can only be stably placed on a
flat surface in a few different ways. This greatly limits the number of possible
placements and thus also the connectivity of the resulting regrasp graph.

\begin{figure}[!t]
\centering
\subfloat[Pot lid organizer]{\includegraphics[width=0.4\linewidth]{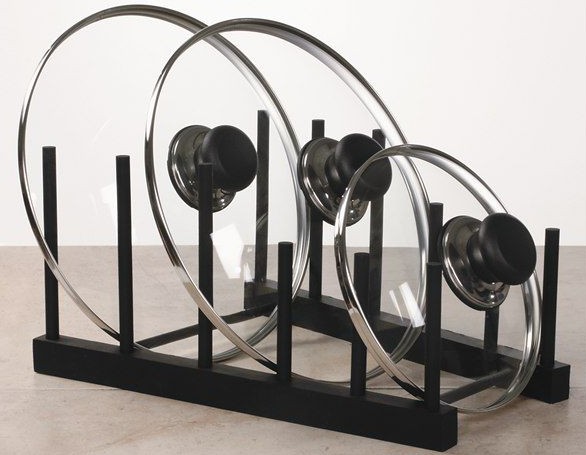}} \qquad
\subfloat[Bicycle kickstand]{\includegraphics[width=0.4\linewidth]{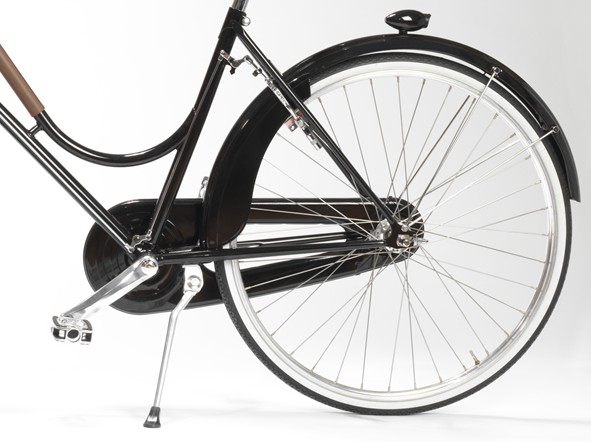}}
\caption{The application of pin supports is popular in everyday life. Ikea's pot
lid organizer uses the pin support to stabilize the pot lids that are otherwise
difficult to be placed. The bicycle kickstand is another example of support
pin which helps to place a bike at a nearly vertical pose.}
\label{fig:motivation}
\end{figure}

To address this challenge, in this paper we use an added pin for the
intermediate placement, instead of only using a flat plane for support. This is
motivated by real world examples of mechanisms that use pins to stabilize items that are otherwise
difficult to orient on a flat surface, such as the pot lid organizer and the bicycle kickstand in
Figure~\ref{fig:motivation}. The main advantage of a support pin lies in its
ability to greatly increase the number of stable placements as well as the
directions from which an object can be manipulated.
{\color{black} Our previous work studying the use of a tiled surface for regrasping (\cite{Wan:ICRA:2015,Wan:CASE:2015}) demonstrated
the benefits of these two properties on the success rate of sequential robotic manipulation. 
Since a pin is able to provide more intermediate states and more candidate regrasps than those through usage of a tilted surface, we expect it to have
better performance while planning the regrasping sequence.}
In particular, we choose one edge $e$ from the object's convex hull, and one
point $x$ from the surface of the object. Then a possible placement can be made
by letting the object touch the pin at point $x$ and touch the flat surface at
edge $e$ (refer to Figure~\ref{fig:solutionflow}(e)). All combinations of $e$
and $x$ correspond to the possible placements, which we then refine to yield valid stable placements through
stability, collision, and friction tests. In this way, we can generate many more stable placements than when
only using a flat plane support, and thus can greatly increase the connectivity
of the regrasp graph. A support pin is also more beneficial for concave
objects since it is able to use the concave part of the objects for the touch point $x$,
while a flat plane would only be able to leverage the convex hull. 

We perform statistical analysis on arbitrary mesh models with thousands of
experiments to demonstrate the advantages of using support pins for regrasp.
Our algorithm automatically computes the stable placements of an object on a
support pin, finds force-closure grasps, generates a graph of regrasp actions,
and searches for regrasp sequences. We use the two-layer regrasp graph
in~\cite{Wan:ICRA:2015} to decouple the search of pick-and-place sequence and
the search of grasps, and delay expensive inverse kinematics and collision
detection computations until necessary. We evaluate the success rates of the
tasks and the length of regrasp sequences with different mesh models, different
pin lengths, and different tasks including reorientation (i.e., flipping) and
assembly. For each task, the added pin is put in various locations relative
to the robot, and the initial and goal poses of the objects are also randomized.
Our results show that an added support pin is beneficial for all tasks. For
some tasks involving pot lid like objects, which are difficult to be reoriented
by a single manipulator through traditional regrasp sequence, the support pin
can significantly increase the success rate of these tasks.

\section{Related Work}
\label{sec:related}
There has been extensive work on techniques related to grasp/regrasp, and of 
these the most relevant are the approaches to object placement planning and sequential
robotic manipulation. Here we give a brief overview of these techniques.

\subsection{Object Placement Planning}
Given an object and a placement area, object placement planning first needs to
figure out the regions that are free for placement. This can be achieved either
according to the environment geometry~\cite{Harada:2014:VOP}, or using a
learning-based framework~\cite{Schuster:2010:Humanoids}. Among the computed free
regions, the next step for object placement planning is to select a placement
location that is suitable for the object, and to determine the object's stable
pose at that location. Most previous approaches only considered placement
locations that are locally flat, and these locations are either
pre-assigned~\cite{Wan:ICRA:2015,Wan:CASE:2015}, or are selected online either
autonomously~\cite{Lozano:2014:IROS} or interactively~\cite{Harada:2014:VOP}. A
few methods~\cite{Jiang:2012:LPN,Baumgartl:2014:ICRA} can leverage non-flat
locations for placement. Jiang~\cite{Jiang:2012:LPN} outlined a learning-based
framework to determine how to place novel objects in complex placement areas,
while Baumgartl et al.~\cite{Baumgartl:2014:ICRA} computed the geometric
information of both the placement area and the object to determine whether a
location is suitable for placement. Once the placement location is determined,
the stable pose of the object at that location is computed by applying the
contact constraint and the stability constraint~\cite{Harada:2014:VOP,Baumgartl:2014:ICRA,Wan:ICRA:2015}. 
Other constraints such as an orientation preference for man-made
objects~\cite{Fu:2008:UOM,Jiang:2012:LPN} can also be taken into account by using
the learning-based framework.
{\color{black}
After a placement has been planned, the robot needs to place the object accurately at the desired pose, which may be challenging in the real-world due to uncertainties. 
To deal with this problem, Kriegman~\cite{Kriegman:1997:LTF} computed a so-called maximal capture region around the desired stable pose in the configuration space; if the object's
initial configuration is within the region, the object is guaranteed to converge to that pose under the force of gravity. Similar ideas were used by~\cite{Moll:2002:MPD} to design the shape of the support surface.
}

In this paper, we focus on designing
the placement area for maximizing the number of stable placements. In
particular, we use a planar surface with one added pin as the placement
area, and thus resulting in many more stable placements than when using only a flat surface or a
surface with complex shapes.

\subsection{Sequential Robotic Manipulation}
Solving manipulation problems requires planning a coordinated sequence of
motions that involves picking and placing, as well as moving through the free
space. Such sequential manipulation problem is challenging due to its high
dimensionality. Early work in this area used an explicit graph search to find a
sequence of regrasping motions~\cite{Lozano-Perez:1992:HRT}. Most recent
approaches are constraint-based. They first formalize the geometric constraints
being involved in the manipulation process, e.g., the object must be in a stable
pose after being placed down, the relative pose between the manipulator and the
object must be fixed during the grasp, and two objects should not collide with
each other. Next, they compute a grasp sequence that can satisfy all these
constraints. Some methods~\cite{Simeon:2004:IJRR,Hauser:2011:IJRR} used these
constraints to define a set of interconnected sub-manifolds in the task space,
and then computed a solution sequence using probabilistic roadmaps embedded in
the constrained space. Other approaches~\cite{Lagriffoul:2012:IROS,Lozano:2014:IROS,Dogar:2015:ICRA} used
these constraints to represent sequential manipulation problems as
constraint-satisfaction problem (CSP), and then solved the CSP using variants of
the backtracking search.

{\color{black}
In order to improve the manipulation flexibility, some recent work began to incorporate non-grasp polices in the manipulation.  
There are many prehensile or non-prehensile strategies besides grasping, such as non-prehensile pivoting~\cite{Carlisle:1994:PGF} and prehensile pivoting~\cite{Yoshida:2010:PBM}, non-prehensile pushing~\cite{Lynch:1996:SPM} and prehensile pushing~\cite{Chavan:2015:PPI}. Regrasp policies leveraging external forces or external environments have also been proposed~\cite{Chavan:2014:EDI,Chavan:2015:PPI} for in-hand manipulation. To make use of these non-grasp policies, Lee et al.~\cite{Lee:2015:HPM} proposed the concept of extended transit, i.e., the transition motion will not only include the transitions between prehensile grasps, but also those between non-prehensile manipulation strategies. A similar idea was also proposed by~\cite{Jentzsch:2015:MOPL}, which used grasping and pushing for transition motion, but only used grasping for transfer motion. 

Multi-arm grasping/regrasping is another active area in sequential robotic manipulation, where the difficulty lies in the high dimensional configuration space of the multi-arm system and the combinatorial complexity due to large number of regrasps and handoffs. Following the initial study by Koga et al.~\cite{Koga:1994:OMA}, sampling-based approaches~\cite{Vahrenkamp:2009:HMP} and grid based searches~\cite{Cohen:2012:SPD} were proposed for planning the manipulation sequence for a dual-arm robot. Both these methods have recently been extended to handle multi-arm manipulation problems~\cite{Dobson:2015:PRA,Cohen:2015:PSM}.

One well-known dilemma for the sequential manipulation is the size of the pre-defined grasp set: a large grasp set will make the planning algorithm too slow, but a small grasp set will result in a high failure rate. Previous approaches tend to use a grasp set with size smaller than 30, e.g., 15 in~\cite{Vahrenkamp:2009:HMP} and 12-14 in~\cite{Cohen:2015:PSM}, and thus they usually are not robust for practical applications. To deal with this challenge, our previous work~\cite{Wan:ICRA:2015,Wan:CASE:2015} proposed a two-layer regrasp graph to leverage a large number of grasps efficiently. 
}
In this paper, we concentrate on generating motion sequences for pick-and-place regrasp
using the regrasp graph proposed in~\cite{Wan:ICRA:2015,Wan:CASE:2015}, which is
applicable to complicated mesh models and large-scale experiments.

\section{Pick-and-place Regrasp using a Support Pin}
\label{sec:algorithm}
In this section, we discuss the details about our pick-and-place regrasp
leveraging a support pin for placement. Our method mainly consists of three
parts: 1) computing all possible stable placements with collision-free grasps
associated; 2) building a regrasp graph whose connectivity reflects the number
of common grasps associated with each pair of different placements; 3) searching
in the regrasp graph for a shortest path between the initial and goal
placements, in order to generate a possible pick-and-place grasp sequence.

\subsection{Placement and Grasp Computation}
We first discuss how to generate all possible placements and the associated
grasps, given the mesh model of an object and the length of the support pin. 
{\color{black} This is the key problem when performing sequential robotic manipulation with a support pin, and has not been studied in contemporary literature. }

\begin{figure*}[!t]
\centering
\includegraphics[width=\linewidth]{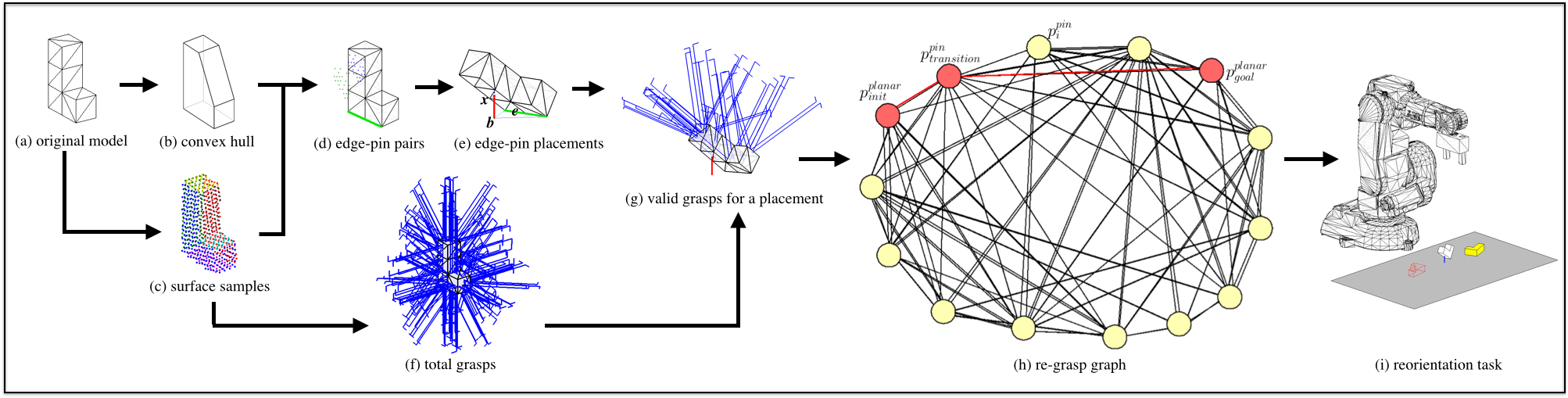}
\caption{The flowchart for computing all possible placements of an object on a
support pin, as well as the grasps associated with each placement. Given an
object in (a), we first compute its convex hull (b), and also perform uniform
sampling on the surface of the object (c). Next, we find all combinations of
convex hull edges and surface samples, and each pair of edge-sample will produce
one edge-pin pair (d), which corresponds to a candidate placement. We only keep
the candidate placements that have passed the stability checking and collision
checking, and then transform them to the world coordinate system to obtain
valid edge-pin placements (e). We also compute the total grasps for the object
(f). Finally, for each placement, we filter out those grasps that are infeasible
due to collision or torque limits, and obtain the valid grasps associated with
each placement (g). These placements and associated grasps are then used to
build a regrasp graph (h), which is used for the reorientation task in (i). The
green edge $e$ in (d) (e) is one edge of the convex hull, and it makes a
placement together with the pin $\overline{bx}$ in (e), where the point $x$ is
one sample on the object surface and the point $b$ is the pin base on the
support plane. The blue segments in (f) and (g) denote grasps, where the longer
segments together with the shorter segments and their ends illustrate the
approaching and opening directions of the grippers, respectively. }
\label{fig:solutionflow}
\end{figure*}

Our method is based on the fact that any placement requires three linearly-independent pivot points. 
Since the support pin already provides one pivot point, the object itself only
needs to provide another two, which correspond to one edge on the object's
convex hull. As a result, we compute placements by finding all combinations of
a single edge on the convex hull and a single point on the object where the pin touches.
Naturally, some of the edge-point pairs may not result in a valid placement due
to collisions or instability, and thus we will need to filter out these invalid edge-point
combinations. An overview of the entire pipeline for placement computation is
shown in Figure~\ref{fig:solutionflow}, and we proceed with a detailed explanation of this pipeline.

\subsubsection{Touching Point Sampling for the Support Pin}
The support pin's candidate touching points on the object are generated by
sampling the mesh surface, as shown in Figure~\ref{fig:solutionflow}(c). To
avoid missing important placement candidates, we want these samples to be evenly
distributed over the object's surface. To achieve this, we first perform
principal component analysis on each face of the object's mesh model to obtain
two main axes for each face. Sample points denoted by $\{x\}$ are then generated uniformly
along these two axes with a reasonable step size. Marginal points too close ($< 5$mm in all our experiments) to
the model's face edges are removed after the sampling, since these points will
not result in a reliable placement. In parallel, we also compute the convex hull
of the object to obtain the candidate edge $e$ where the object touches the flat
support plane, as shown by the green line in Figure~\ref{fig:solutionflow}(d).

\subsubsection{Finding Base Point of the Support Pin}
Given a sample touching point $x$, and one edge $e = \overline{e_1 e_2}$ on the
convex hull, we need to determine whether the $(x,e)$ pair will make a valid
placement on a pin with length $l$. If a valid placement exists, we need to then compute the
orientation of the pin relative to the object. For this purpose, we first
compute the distance from $x$ to a line passing through the endpoints $e_1$ and
$e_2$. If such distance is shorter than $l$, then it is impossible to find a
valid placement. Otherwise, we continue to find a point $b$ at which the pin touches the support plane and
satisfies the following constraints:
\begin{align}
\left \{ \begin{array}{c} \| x - b\| = l \\
(x - b) \cdot (e_1 - b) = 0 \\ 
(x - b) \cdot (e_2 - b) = 0
\end{array}
\right.
\end{align}
Solving these three equations yields two solutions for $b$, one of which can be easily discarded. 
The edge $e$ and the pin $\overline{bx}$ together form one candidate placement. We need to
verify the validity of this candidate placement by checking whether it
satisfies a set of constraints. First, the pin $\overline{bx}$ should not
collide with the object. Second, this placement should be stable, i.e., the
object's center of mass should have a projection inside the triangle $\Delta e_1
e_2 b$. Finally, the angles between the pin direction and the touching face's
normal should be within a range determined by a given friction factor $\mu$, to
make sure that the object will not slide over the pin. We only keep the
candidate placements that satisfy all these constraints. {\color{black} For each convex hull
edge, there may be more than one associated placement, and we choose the placements
that result in the highest center of mass for the object. This is because for an object
with uniform distribution of mass, a higher center of mass means that more space will be left
for collision-free grasps, which increases the connectivity of the regrasp graph.}
The collection of all placements is denoted as $\mathbf p^{\text{pin}} =
\{p_1^{\text{pin}}, p_2^{\text{pin}}, ...\}$. In our experiment, we will also
use the traditional placements only leveraging a planar surface as the
support~\cite{Wan:ICRA:2015}, which are denoted as $\mathbf p^{\text{planar}}$.

\subsubsection{Calculating Total Grasps}
\label{sec:algo:totalgrasp}
We then compute the set of total grasps, namely $G = \{g_1, g_2. ...\}$,
associated with the object without considering collision-free and inverse
kinematics constraints. One example of the total grasps is shown in
Figure~\ref{fig:solutionflow}(f). Each grasp $g_i$ is computed according to the
algorithm in~\cite{Wan:ICRA:2015} such that it satisfies force-closure
constraints and consists of the position and orientation of the robot hand. For
the Robotiq 2-finger adaptive gripper that we used, the grasps are computed by
first examining possible parallel face pairs on the object, and then sampling
the rotation direction around normals of the parallel faces. We can adjust the
number of sampled directions to control the grasp density. We will discuss the
relationship between the grasp density and the success rates of the
pick-and-place tasks in Section~\ref{sec:experiment}.

\subsubsection{Grasps Associated with Placements}
Possible grasps associated with a pin-based placement are found by transforming
the total grasps from the object's local coordinate to the current world frame
of the placement, and then checking the collision between the gripper and
the support pin, as well as the collision between the object and the flat support surface. After filtering out
the invalid grasps, we obtain all valid grasps associated with a
placement, and we denote the association as $\{p_i, G^i\}$,
where $p_i$ is a placement from either $\mathbf p^{\text{pin}}$ or $\mathbf
p^{\text{planar}}$ and $G^i = \{g_p^i, g_q^i, ...\}$ is the valid grasp set for $p_i$. 
In Figure~\ref{fig:lPinVSPlane}(b), we show all the placements and associated grasps
for one ``L'' shaped object (as shown in Figure~\subref*{fig:model:1t1p}) using the
support pin for placement. For comparison, we also use the method
in~\cite{Wan:ICRA:2015} to compute the placements and grasps for the traditional
regrasp of placing objects only on a flat surface, and the results are shown in
Figure~\ref{fig:lPinVSPlane}(a).

{\color{black}
\subsection{Regrasp Graph Construction}
Given the placements and the associated grasps, we build a two-layer regrasp
graph where the first layer is composed of placements and the second layer is composed of grasps.
The hierarchical structure of the regrasp graph can help to delay the computation for inverse kinematics and collision checking until necessary,
and thus reduces the combinatorial complexity of the manipulation planning.

In the first layer, two placements $p_i$ and $p_j$ are connected by one edge if their associated grasp sets
$G^i$ and $G^j$ are not disjoint, i.e., $G^i \cap G^j \neq \emptyset$. 
In the second layer, we add edges between grasps that are shared by two placements. For instance, given two connected placements $p_i$ and $p_j$, suppose the intersection of their associated grasp sets is $G^i \cap G^j = \{g_u, g_v, ...\}$. Then we will add edges $(g_u^i, g_v^i)$, $(g_u^j, g_v^j)$, ..., and $(g_u^i, g_u^j)$, $(g_v^i, g_v^j)$ in the second layer graph, where we use $g_*^i$ and $g_*^j$ to denote $g_*$'s corresponding grasps in $G_i$ and $G_j$, respectively. 
For more details about how to build this graph, please refer
to~\cite{Wan:ICRA:2015}. Examples of the first layer of the regrasp graph are
shown in Figure~\ref{fig:solutionflow}(h) and Figure~\ref{fig:regraspgraph}.

\subsection{Grasp Sequence Computation}
The motion sequence for manipulating the object to perform reorientation or assembly
tasks can be generated by first searching in the first layer regrasp graph for a shortest
path connecting the initial and goal placements (as shown by the
$p_{\text{init}}^{\text{planar}}$ and $p_{\text{goal}}^{\text{planar}}$ in
Figure~\ref{fig:solutionflow}(h)). Then in the second layer, by considering the connectivity 
between the grasps associated with the placements along the shortest path, a sequence of grasps
can be generated. After that, we perform collision checking and solve for
inverse kinematics at each pick-up and place-down moment.
One example of the generated grasp sequence is shown in Figure~\ref{fig:grasp_sequence}.
}

\section{Experiments and Analysis}
\label{sec:experiment}
In this section, we demonstrate the advantage of using an added support pin
for intermediate placement in the pick-and-place regrasp. In particular, we
perform a large number of reorientation tasks in a simulated environment. We
compare the success rates and the lengths of the resulting regrasp sequences in
the presence of two different placement settings: one only using the flat plane,
and the other using an added support pin. Meanwhile, the effects of
different grasp density and pin length are compared against the above two
criteria. We also use the support pin placement to accomplish one assembly task
which is not feasible while only using planar surface as support.

{\color{black}
We implemented our algorithms in Matlab on an Intel Core i7 CPU
running at 3.40GHz with 32GB of RAM and running Ubuntu 12.04 LTS. All the timing results are generated using a single core.
}

\subsection{Experiment Settings}
\begin{figure}[!t]
\centering
\subfloat[$l$]{\includegraphics[height=1.3cm]{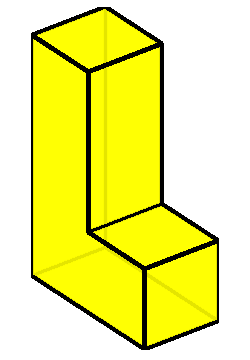}} 
\subfloat[$el$]{\includegraphics[height=1.3cm]{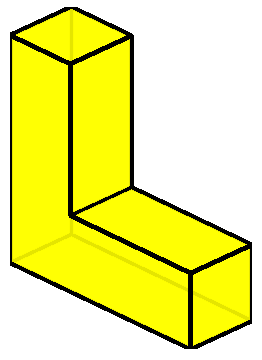}} 
\subfloat[$3t$]{\includegraphics[height=1.3cm]{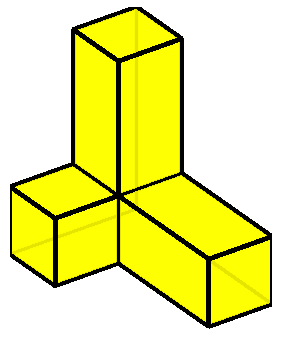}} 
\subfloat[$3ts$]{\includegraphics[height=1.3cm]{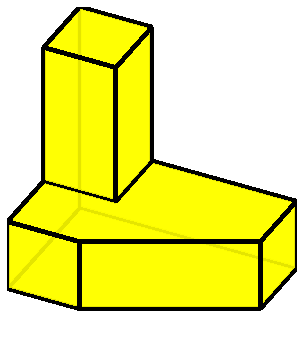}} 
\subfloat[$3t2s$]{\includegraphics[height=1.3cm]{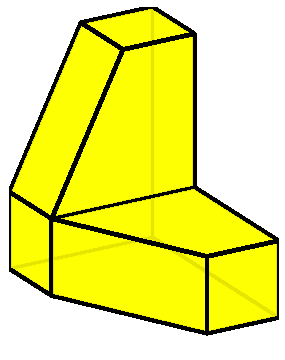}}
\subfloat[$cross$]{\includegraphics[height=1.3cm]{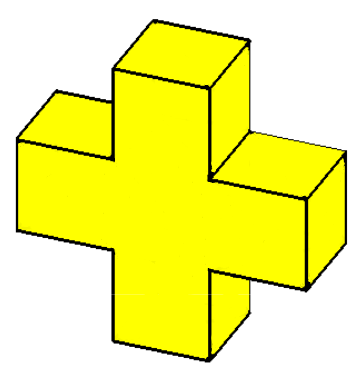}} 
\subfloat[$t$]{\includegraphics[height=1.3cm]{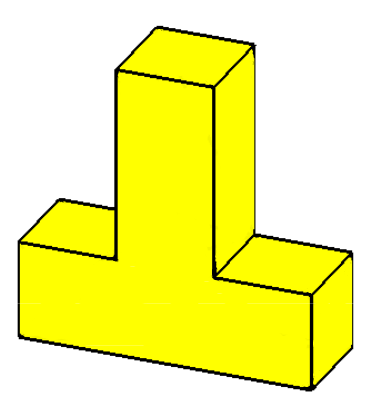}} 
\subfloat[$1t1p$]{\includegraphics[height=1.3cm]{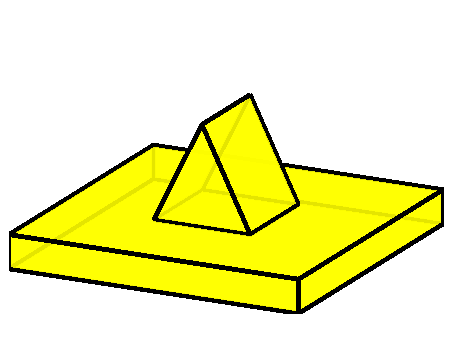} \label{fig:model:1t1p}}
\caption{Mesh models for all objects used in our experiments. All objects are non-convex, and $1t1p$ is an object of the pot lid shape. 
{\color{black} The first five objects (a)-(e) form a morphing sequence where we increase the object's volume by gradually adding more elements to the shape $l$. 
The height of all objects from (a) - (g) is $9$ cm, and all objects can be bounded in a cube of size $9$ by $9$ by $9$ cm.}}
\label{fig:model}
\end{figure}

\begin{figure}[!t]
\centering
\includegraphics[width=0.5\linewidth]{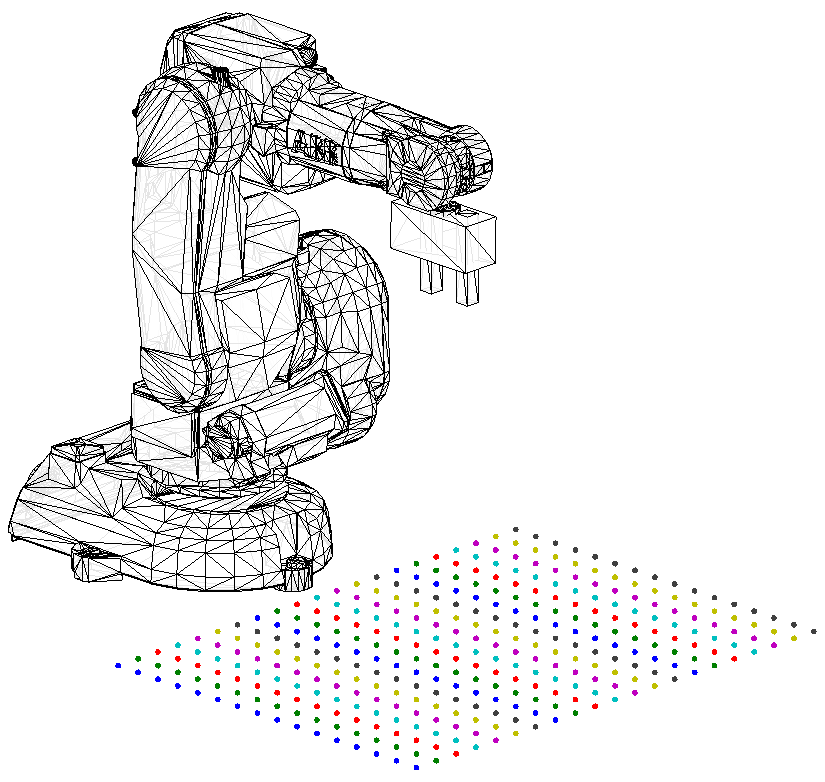}
\caption{The experimental setting for the reorientation task. The $0.8$ by $0.6$ meter rectangular working area in front of an ABB IRB140 robot
 is evenly divided into a $20$ by $15$ grid of $4$ by $4$ cm grid squares.}
\label{fig:experimentsetting}
\end{figure}

\begin{figure*}[!t]
\centering
\subfloat[]{\includegraphics[width=0.15\linewidth]{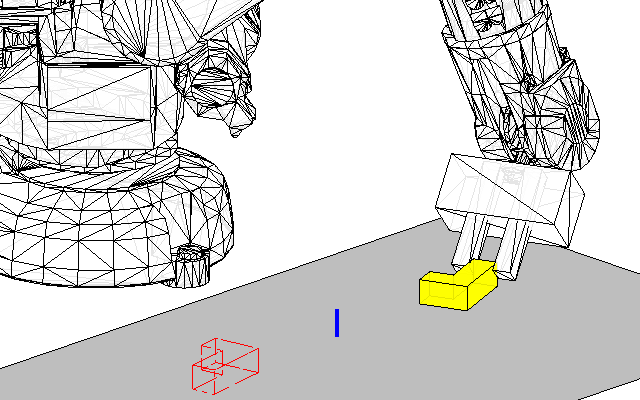}} 
\subfloat[]{\includegraphics[width=0.15\linewidth]{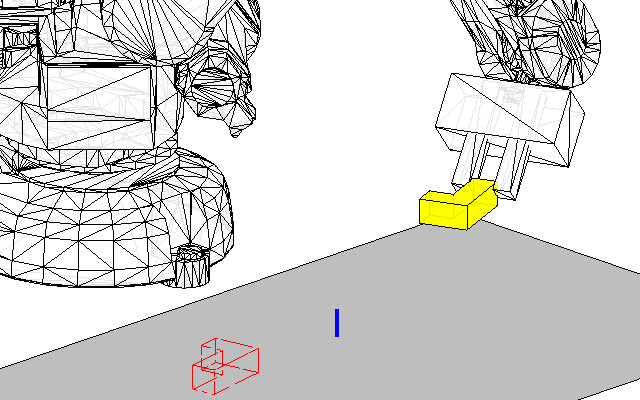}} 
\subfloat[]{\includegraphics[width=0.15\linewidth]{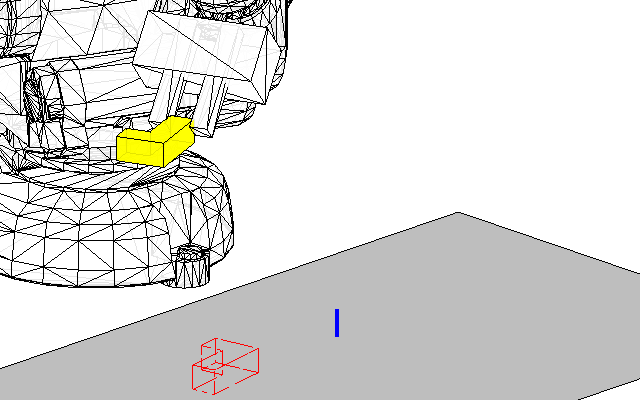}} 
\subfloat[]{\includegraphics[width=0.15\linewidth]{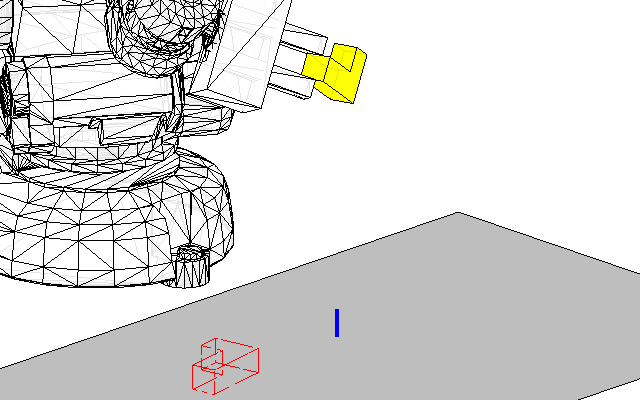}}
\subfloat[]{\includegraphics[width=0.15\linewidth]{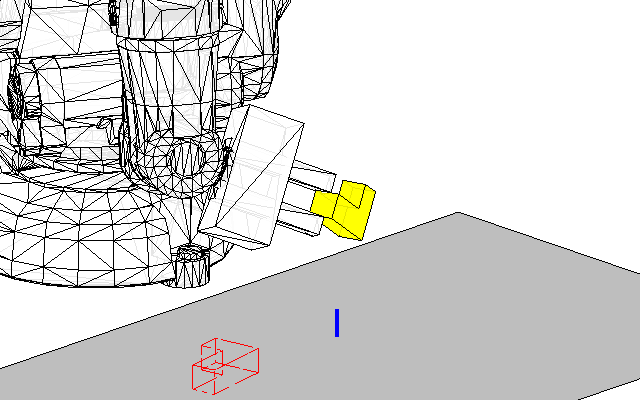}}  
\subfloat[]{\includegraphics[width=0.15\linewidth]{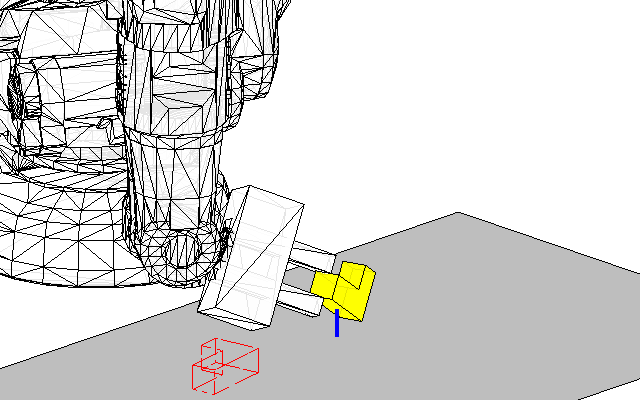}} \\
\subfloat[]{\includegraphics[width=0.15\linewidth]{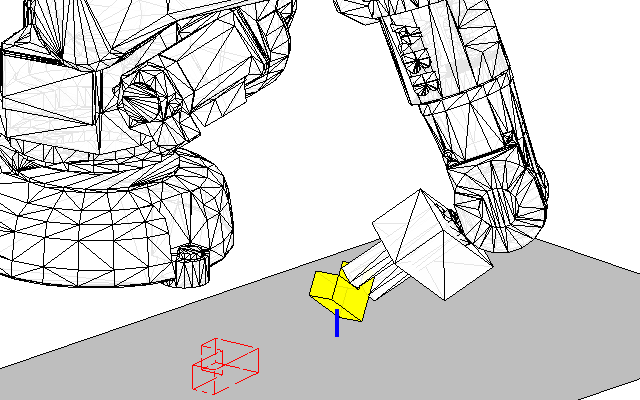}} 
\subfloat[]{\includegraphics[width=0.15\linewidth]{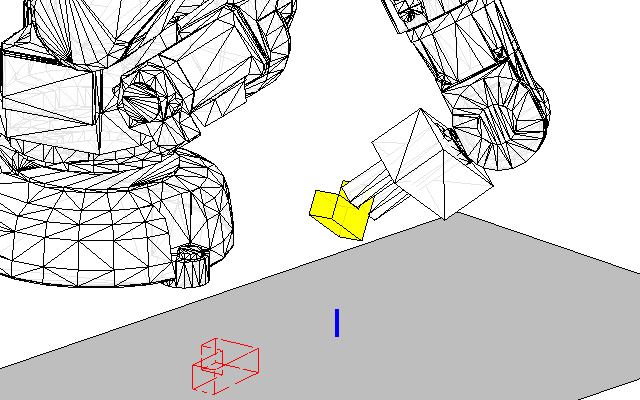}} 
\subfloat[]{\includegraphics[width=0.15\linewidth]{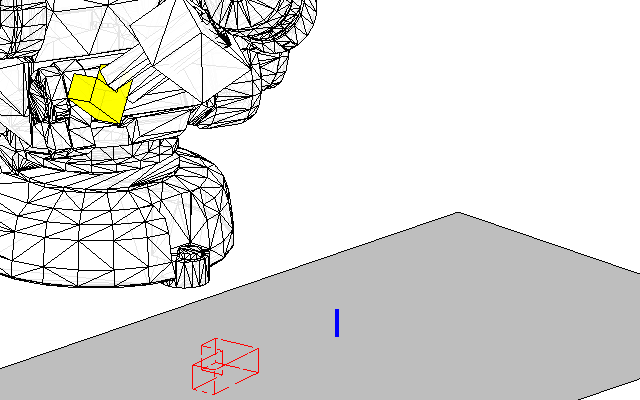}} 
\subfloat[]{\includegraphics[width=0.15\linewidth]{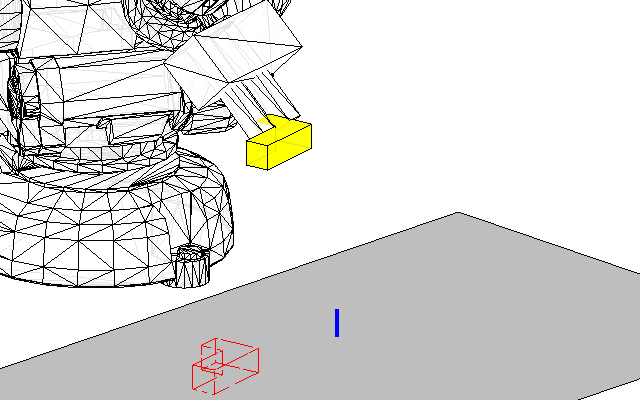}} 
\subfloat[]{\includegraphics[width=0.15\linewidth]{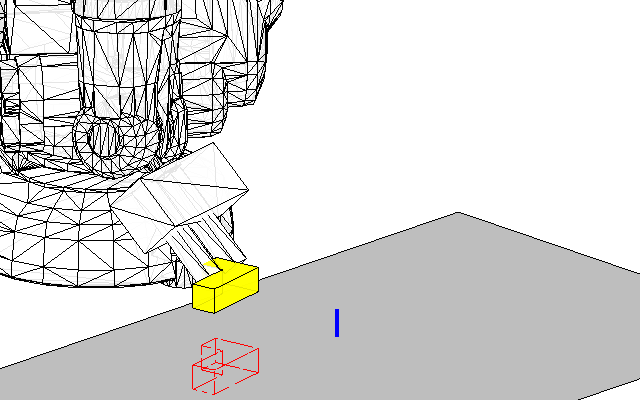}} 
\subfloat[]{\includegraphics[width=0.15\linewidth]{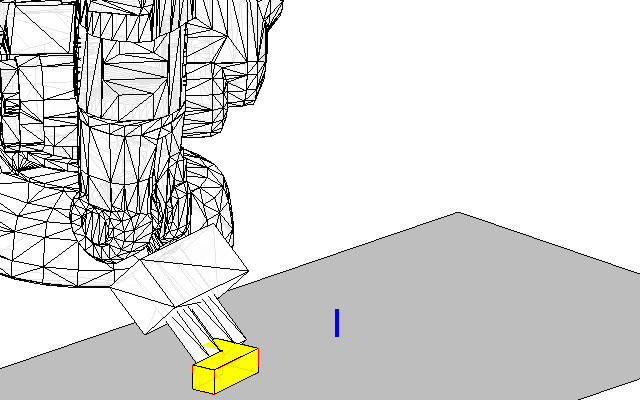}} 
\caption{The regrasp sequence for reorienting an ``L'' shaped non-convex model using one support pin as intermediate placement. The sequence is computed with the two-layer regrasp graph built on the grasps computed by sampling the mesh model. (f) and (g) are the transition steps while regrasps occur. }
\label{fig:grasp_sequence}
\end{figure*}

We use an ABB IRB140 robot manipulator with a 2-finger Robotiq gripper 85
mounted as the end-effector to repeatedly perform reorientation tasks for
several different object models as illustrated in Figure~\ref{fig:model}. Among these object models,
(a)-(e) form a morphing sequence where we increase the object's volume by gradually adding more elements to the shape $l$.
We select a $0.8$ by $0.6$ meter rectangular working area in front of
the robot, as shown in Figure~\ref{fig:experimentsetting}. This working area is
evenly divided into a $20$ by $15$ grid of $4$ by $4$ cm grid squares.
We let each corner of each grid square, $x_{\text{grid}}$, be the initial and goal positions
for the object to be manipulated, while the initial and goal orientations are randomized. 
The object's initial and goal placements are also randomly chosen from $\mathbf p^{\text{planar}}$, 
i.e., the beginning and end of the manipulation are always placements on the flat surface.
For regrasp transitions, the object will be placed in an intermediate location which is $20$ cm to the left of
$x_{\text{grid}}$ in order to avoid the collision between the pin and the
object at initial and goal poses. Ten trials were performed for each corner of the grid square, and thus, in total, there are $(20+1) \times (16 + 1)
\times 10 = 3360$ trials for each object. An example of the reorientation
sequence is shown in Figure~\ref{fig:grasp_sequence}, where the robot uses one
regrasp placing on the pin to successfully reorient one ``L'' shaped object.
{\color{black} Unless otherwise stated, the length of the pin used in our experiment is $3$ cm, and the grasp density is $8$ directions. }

\subsection{Comparison of Placements and Grasps}
\label{sec:experiment:placementsAndGrasps}
\begin{figure}[!t]
\centering
\subfloat[Placements and grasps with only a support plane
]{\includegraphics[width=0.7\linewidth]{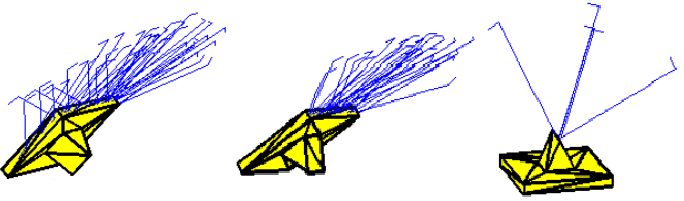}} \\
\subfloat[Placements and grasps with an added support
pin]{\includegraphics[width=\linewidth]{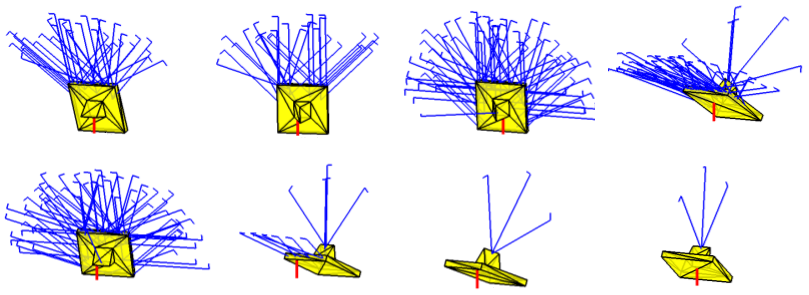}}
\caption{For the pot lid like object $1t1p$, we compare its placements and associated grasps between only using the planar surface as support and using an
added support pin as the placement area. }
\label{fig:1t1pPinVSPlane}
\end{figure}

\begin{figure}[!t]
\centering
\subfloat[Placements and grasps with only a support plane]{\includegraphics[width=0.7\linewidth]{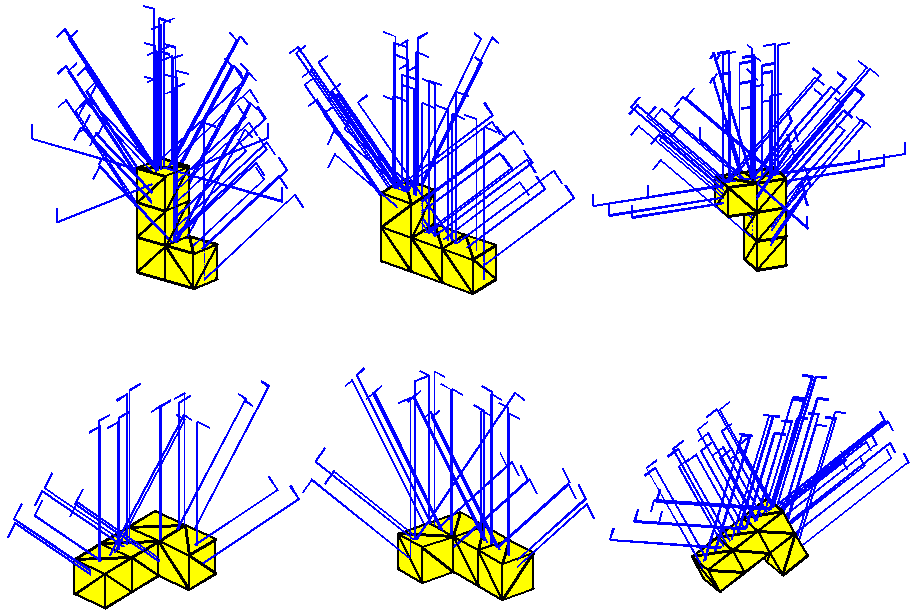}} \\
\subfloat[Placements and grasps with an added support
pin]{\includegraphics[width=0.99\linewidth]{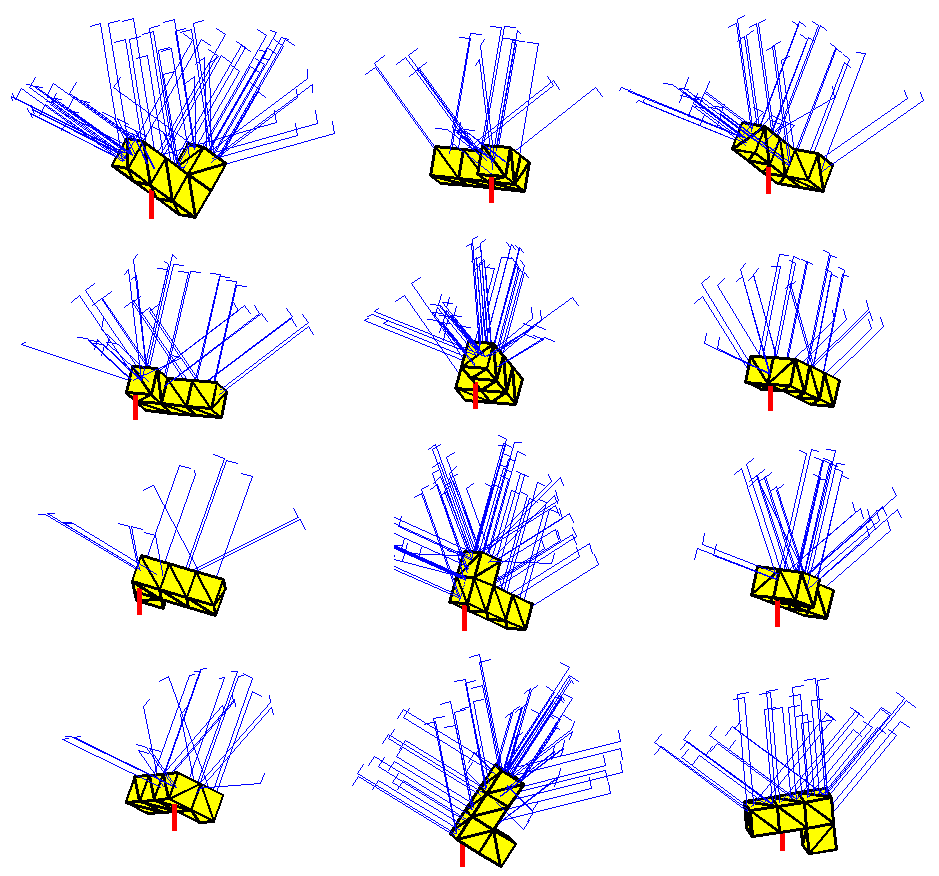}}
\caption{For the ``L'' shaped object $l$, we compare its placements and associated grasps between only using the planar surface as support and using an
added support pin as the placement area.}
\label{fig:lPinVSPlane}
\end{figure}

We first compare the cases with only planar support to the cases with an added support pin
in terms of placements and associated grasps. In Figure~\ref{fig:1t1pPinVSPlane} and Figure~\ref{fig:lPinVSPlane}
we show the results for a pot lid like object $1t1p$ and for an ``L'' shaped
object $l$.

\begin{figure}[!t]
\centering
\subfloat[]{\includegraphics[width=0.45\linewidth]{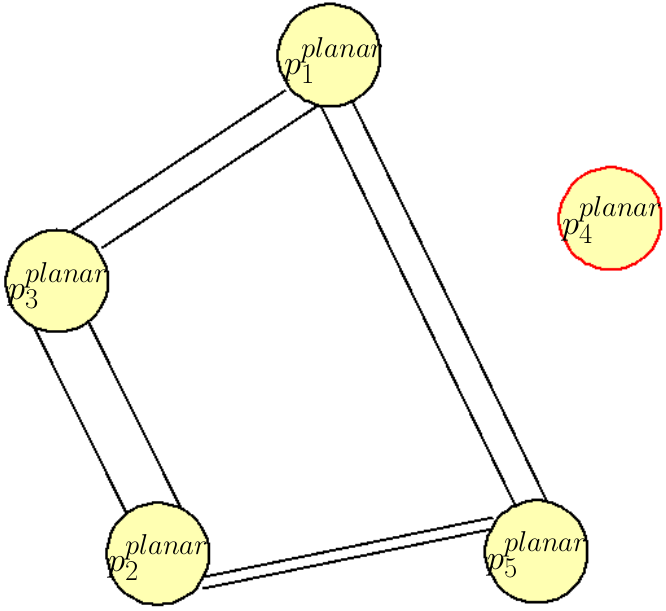}} \qquad
\subfloat[]{\includegraphics[width=0.45\linewidth]{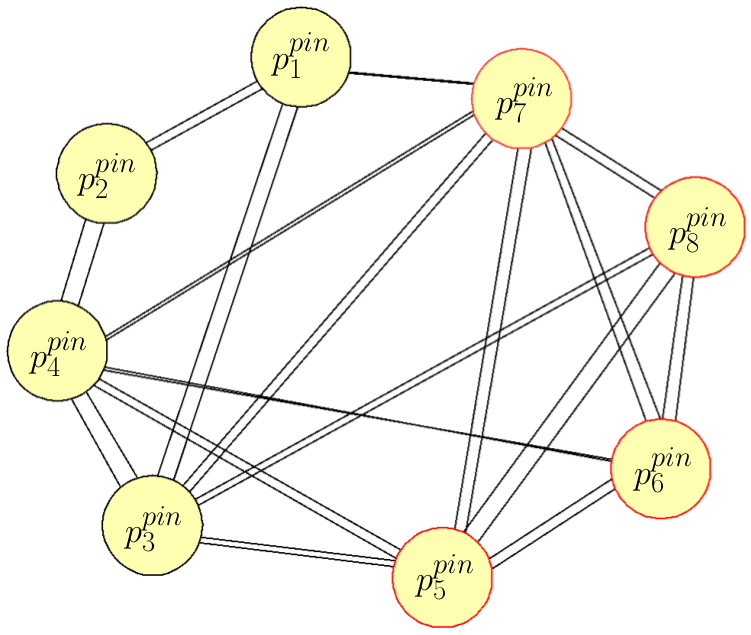}} 
\caption{The regrasp graphs of $1t1p$ for the planar placement and for the pin
placement, respectively. (a) is the regrasp graph for planar placement, and
$p^{\text{planar}}_4$ is the $4$-th body-down placement in
Figure~\ref{fig:1t1pPinVSPlane}(a), and we can see it has no connection with
other placements in the graph, and this suffers the connectivity of the entire
graph. (b) is the regrasp graph for the pin placement, which has more placements
than the planar support case. In addition, $p^{\text{pin}}_5$,
$p^{\text{pin}}_6$, $p^{\text{pin}}_7$ and $p^{\text{pin}}_8$ correspond to the
body-down placements in Figure~\ref{fig:1t1pPinVSPlane}(b), and they are well
connected to the other placements thanks to the usage of the added support
pin.}
\label{fig:regraspgraph}
\end{figure}

The pot lid shape $1t1p$ has a large body and a small handle, and all stable
placements for it could be roughly categorized into two scenarios: handle-down and
body-down. While only using the planar surface as support, there are only five
different placements as shown in Figure~\ref{fig:1t1pPinVSPlane}. Even worse is
that the grasps associated with the body-down placements are limited only to the
handle part and that there are very few of such grasps, as illustrated by the
placement $p_4^{\text{planar}}$ in Figure~\ref{fig:1t1pPinVSPlane}(a). As a
result, the connectivity is low in the corresponding regrasp graph and sometimes
there may even be no connection between two placements as shown in
Figure~\ref{fig:regraspgraph}(a). With the help of a pin, we note an increase in interconnected
placements added to the graph and also an increase in common grasps 
associated with different placements, as shown in
Figure~\ref{fig:1t1pPinVSPlane}(b). This in turn increases the connectivity of the
regrasp graph, as shown in Figure~\ref{fig:regraspgraph}(b).

As shown in Figure~\ref{fig:lPinVSPlane}, the ``L'' shaped object $l$ also
has more stable placements while using the pin support, which helps to increase the
connectivity of the resulting regrasp graph.

\begin{figure}[!t]
\centering
\subfloat[Placements and grasps with only a support plane]{\includegraphics[width=\linewidth]{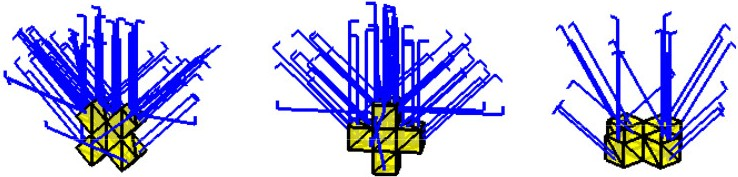}} \\
\subfloat[Placements and grasps with an added support
pin]{\includegraphics[width=0.8\linewidth]{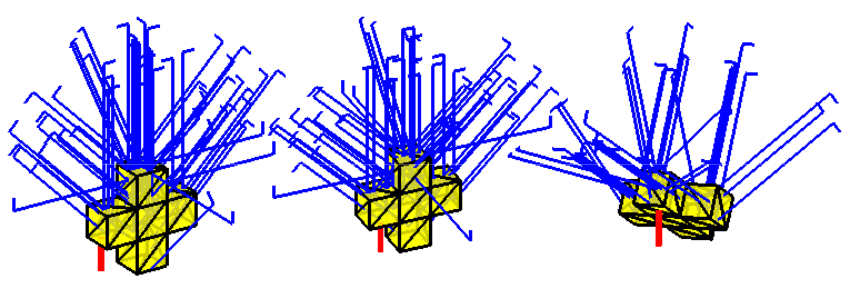}}
\caption{For the $cross$ model, we compare its placements and associated grasps between only using the planar surface as support and using an
added support pin as the placement area.}
\label{fig:crossPinVSPlane}
\end{figure}

\subsection{Comparison of Success Rates}
\label{sec:experiment:avgsuccess}
\begin{figure}[!t]
\centering
\includegraphics[width=1\linewidth]{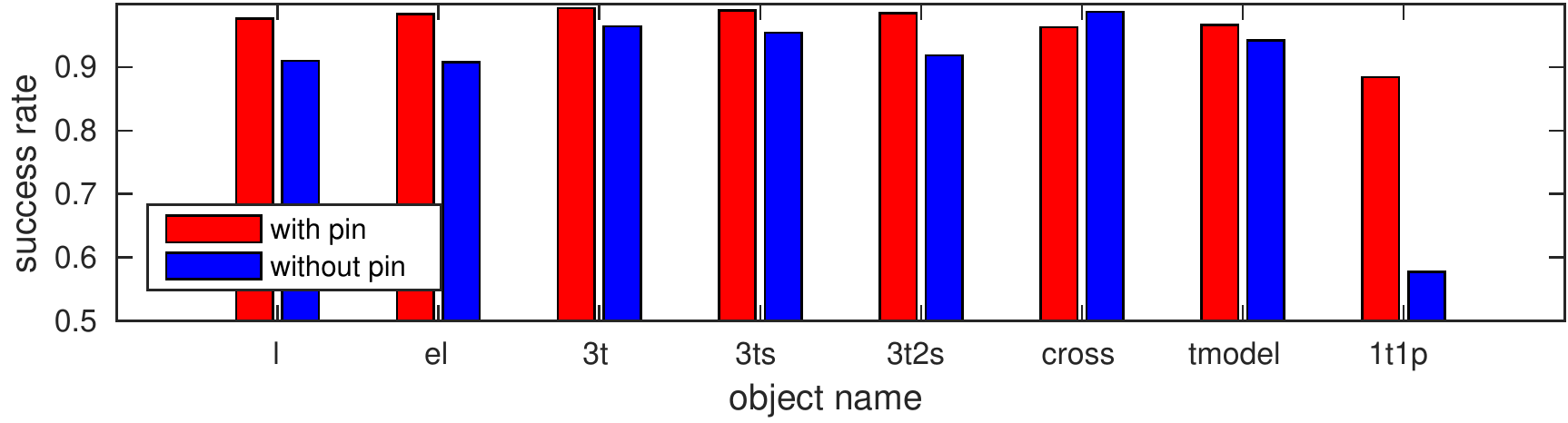}
\caption{Comparison between the average success rates of the orientation task,
while using two different placement settings. The red bars are the results
using the added pin for placement, while the blue bars are the results only
using the planar surface for support.}
\label{fig:successrate}
\end{figure}

We then compare the reorientation task's success rates while using the pin and
planar placement settings. From the result shown in
Figure~\ref{fig:successrate}, we can see that the success rate of the pin
placement is higher than that of the planar placements for all different
objects except the $cross$ model. The advantage of the pin placement over the planar placement is
significant for the object $1t1p$ with the pot lid shape. This is due to the
significant improvement of the regrasp graph's connectivity while using the pin
placement rather than the planar placement, as mentioned above in Section~\ref{sec:experiment:placementsAndGrasps}. For other
objects, the success rate improvement of the pin placement is not as significant
as the pot lid shape object, because their regrasp graph's connectivity is
already rich enough even while only using the planar support. 
{\color{black} For the $cross$ model, the success rate of
the pin placement is lower than that of the planar placement. This is because the $cross$ model already has 
many different placements with a large set of associated grasps while only using the planar support.
The added pin does not bring many more valid grasps; instead, one type of placement (the first in Figure~\ref{fig:crossPinVSPlane}) disappears since it violates
the stability constraint in the case of the pin placement.

We further investigate how the success rate changes along with the deformation of the object's shape, by analyzing the success rate results 
for the morphing sequence (a)-(e) in Figure~\ref{fig:model}. According to Figure~\ref{fig:successrate}, we observe that when more elements are added to the object,
the success rate first increases and then decreases when using the planar placement. This phenomenon can be explained as follows. 
In the beginning, the added elements increase the shape complexity of the object's convex hull, and this leads to a higher success rate. 
After many elements have been added, the shape of the object's convex hull becomes less complicated or more ``round'', and this results in fewer placements and grasps, and eventually a lower success rate. 
One reasonable conjecture is that an object with richer features will have a higher success rate using the planar placement. When using the pin placement, we observe that all objects have similar success rates, because the pin adds a significant feature to all shapes, and this reduces the difference in the shape complexity among different shapes' convex hulls.
}

\subsection{Relations between the Pin Length and Success Rates}
\begin{figure}[!t]
\centering
\subfloat[$l$]{\includegraphics[width=0.93\linewidth]{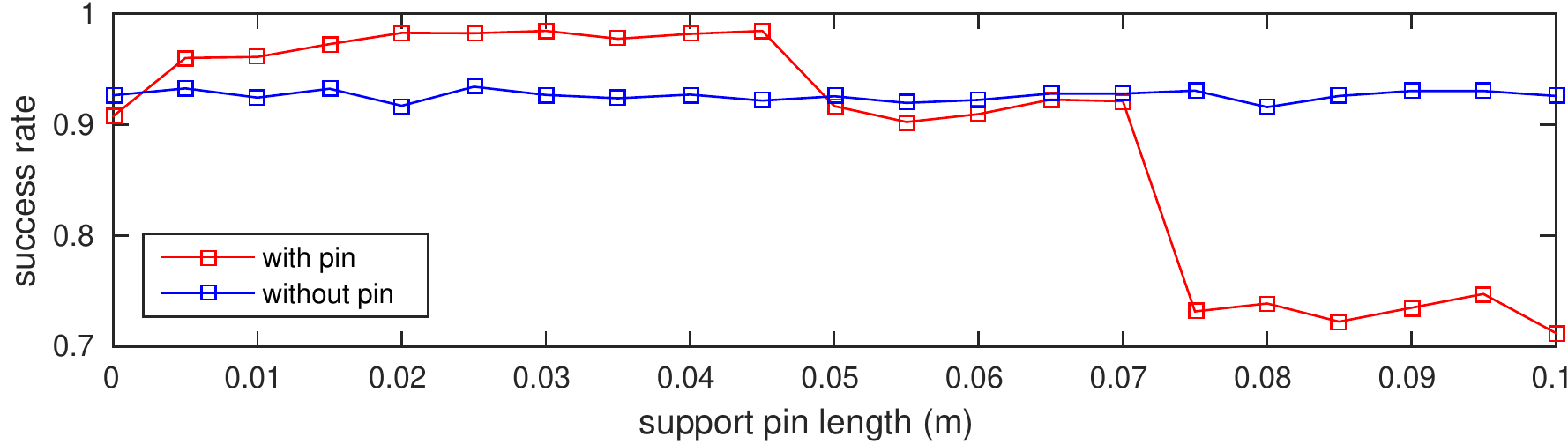}} \vspace*{-0.08in} \\
\subfloat[$el$]{\includegraphics[width=0.93\linewidth]{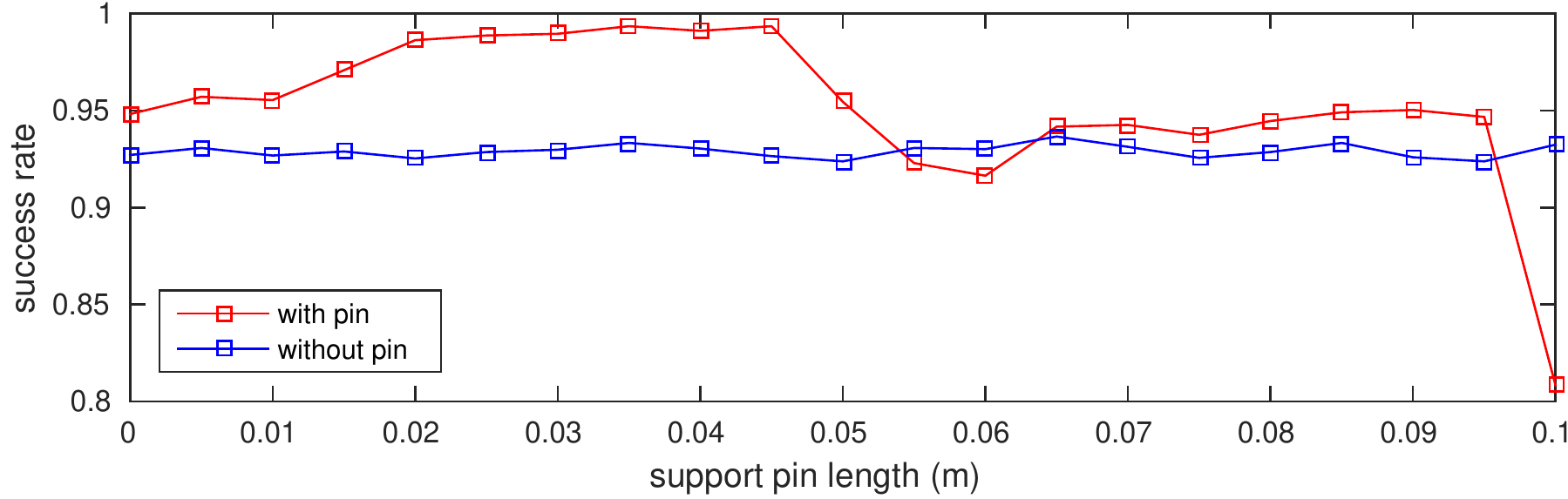}} \vspace*{-0.08in} \\
\subfloat[$3t$]{\includegraphics[width=0.93\linewidth]{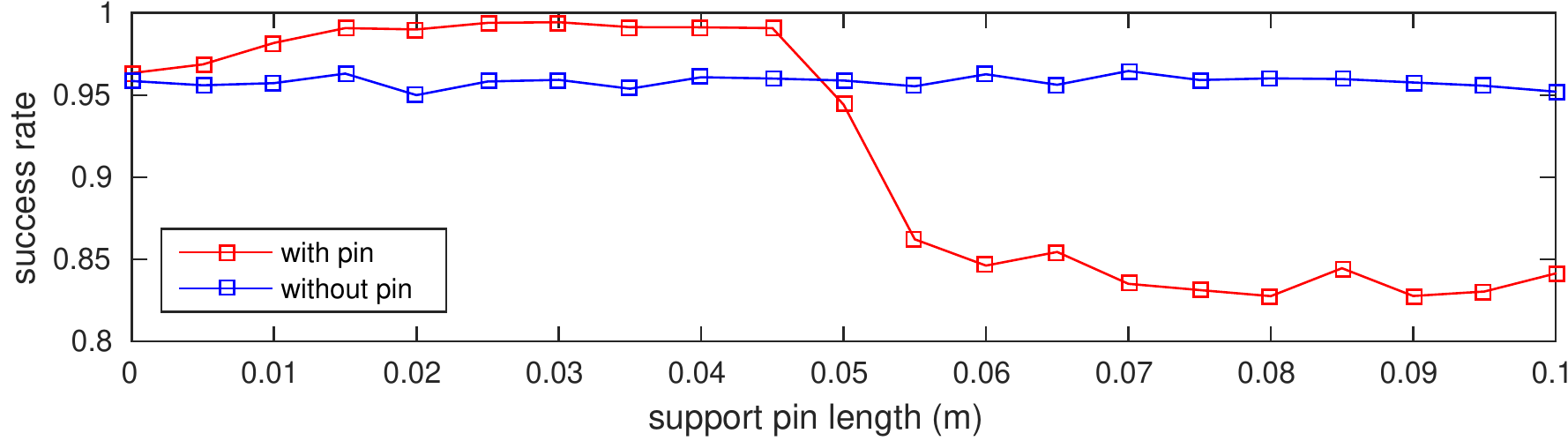}} \vspace*{-0.08in} \\
\subfloat[$3ts$]{\includegraphics[width=0.93\linewidth]{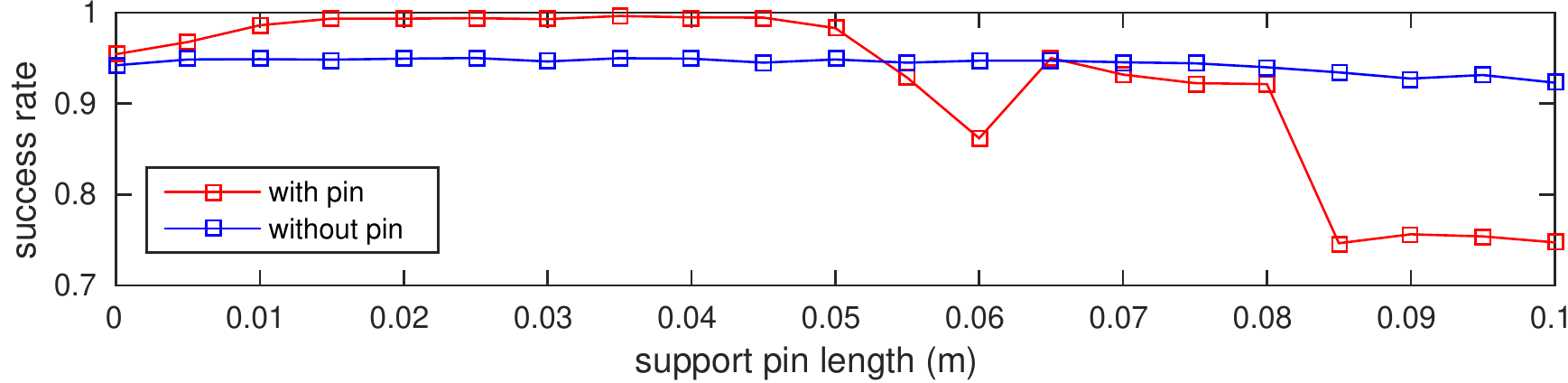}} \vspace*{-0.08in} \\
\subfloat[$3t2s$]{\includegraphics[width=0.93\linewidth]{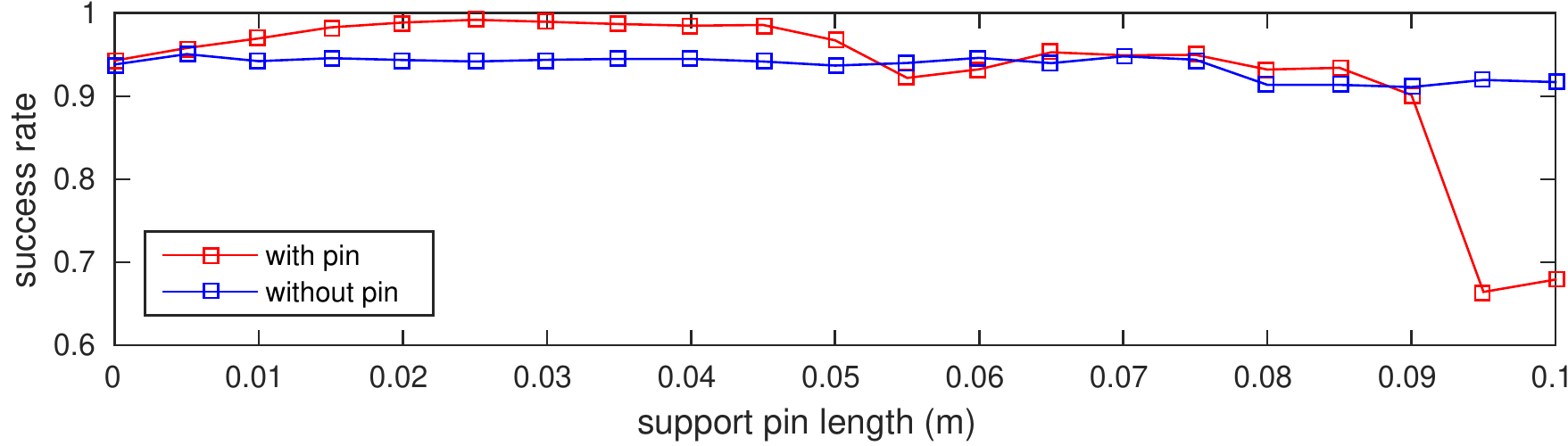}} \vspace*{-0.08in} \\
\subfloat[$cross$]{\includegraphics[width=0.93\linewidth]{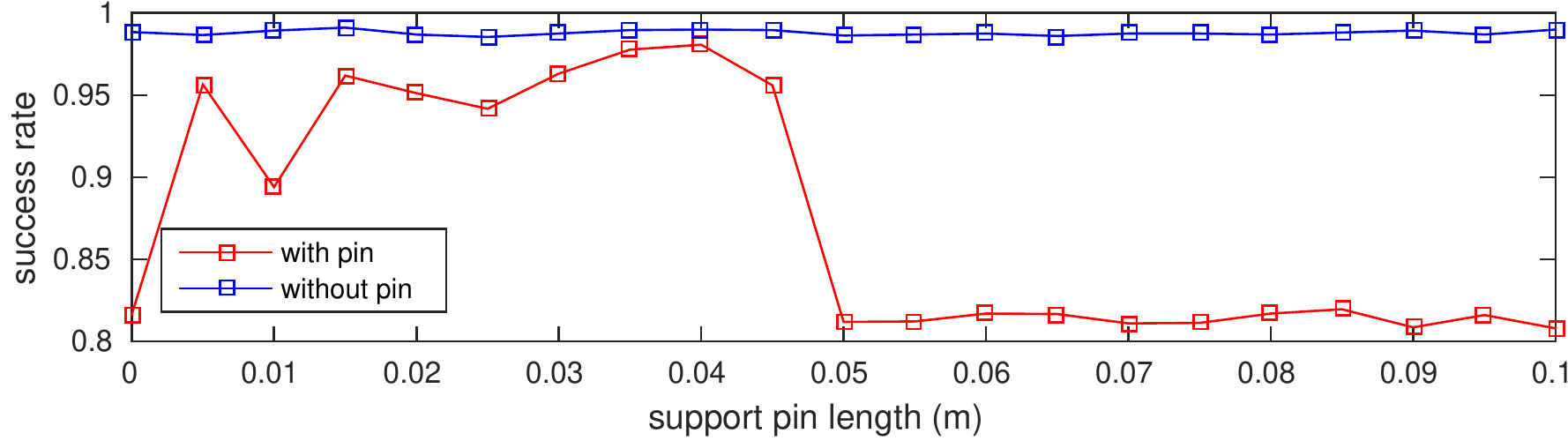}} \vspace*{-0.08in} \\
\subfloat[$t$]{\includegraphics[width=0.93\linewidth]{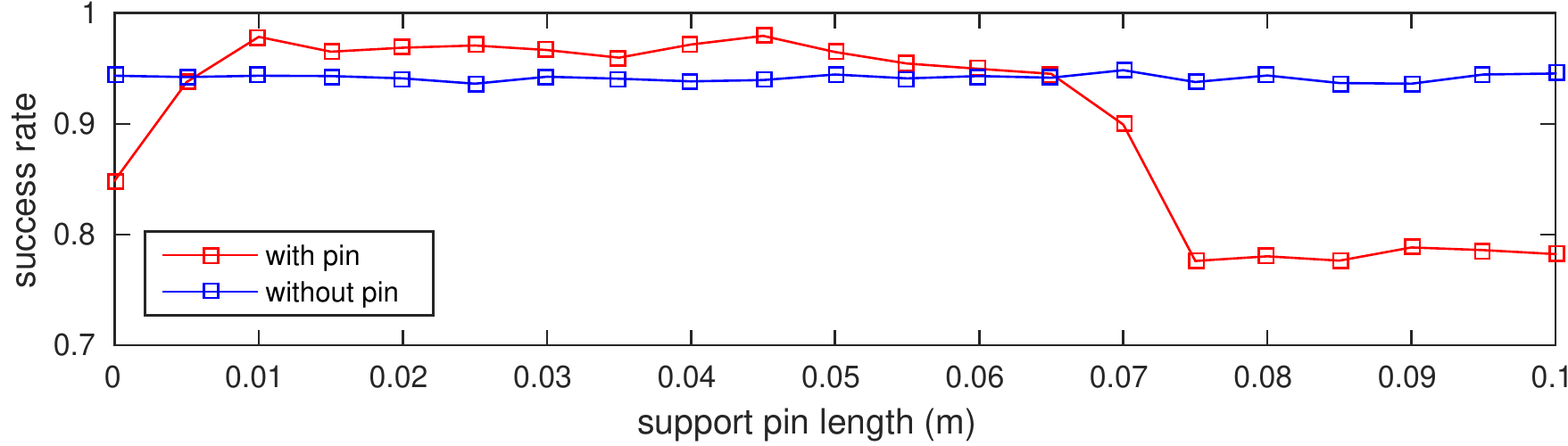}} \vspace*{-0.08in} \\
\subfloat[$1t1p$]{\includegraphics[width=0.93\linewidth]{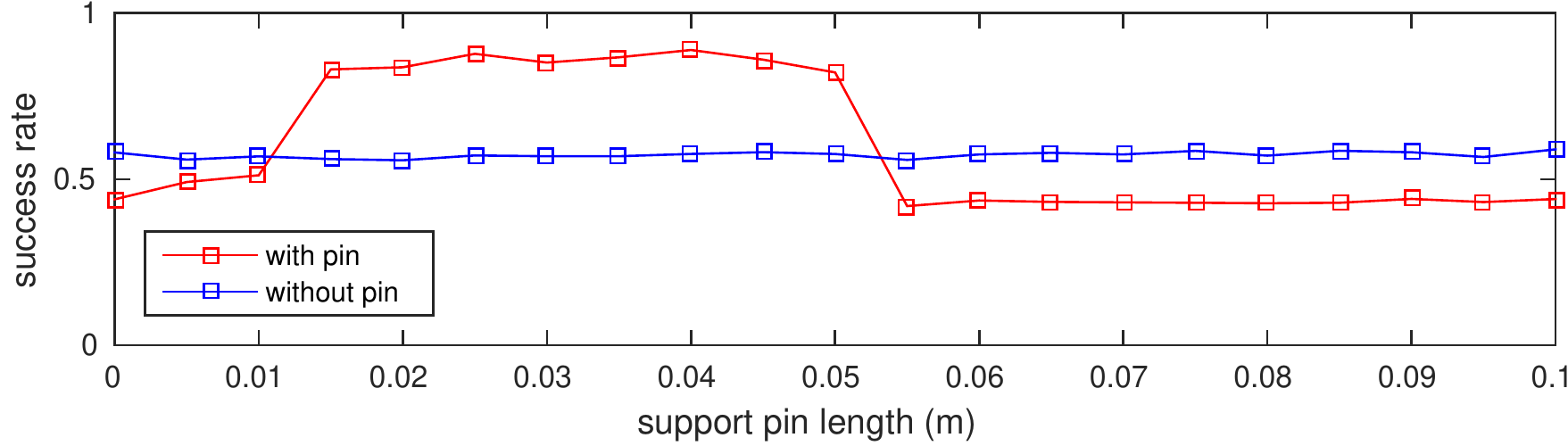} \label{fig:stickLengthResults:1t1p}} 
\caption{Relations between the success rate and the support pin's length while reorienting different models.}
\label{fig:stickLengthResults}
\end{figure}

In this part, we investigate the relationship between the length of the support
pin and the success rate of the reorientation task. 
{\color{black} We change the length of the
pin and repeat the experiments above. The results are shown in Figure~\ref{fig:stickLengthResults}.
We can observe that when the pin length is between 2cm and 4cm, the success rate of the pin placement is
higher than that of the planar placement for all objects except the $cross$ model. However, when the pin length is too long or too short, the success rate
of the pin placement may decrease, because the object can not be stably placed on the pin. 

In particular, for the pot lid shape object
$1t1p$ as shown in Figure~\subref*{fig:stickLengthResults:1t1p}, the success rate of the pin
placement can be even lower than the planar placement when the pin's length is shorter than $1$ cm. This is because for a very short
pin, many candidate placements might fail to satisfy the stability, collision
and friction constraints simultaneously. In particular, when the handle-side is
down, a short pin will result in the object colliding with the planar surface, or
will result in a relative angle with the object that is larger than the friction
angle so that the object will slide over the pin. Thus the number of valid
placements on the pin might be even fewer than that of the planar placement
setting, and this in turn results in a regrasp graph with lower connectivity.
When the pin length is longer than $1$ cm, the successful rate ``jumps" to around $85\%$ and remains stable thereafter, which is better than that of the the planar placement. 

For the $cross$ model, the success rate of the pin placement is always lower than the planar placement, and the reason is similar to what has been explained in Section~\ref{sec:experiment:avgsuccess}. But still we can observe that the success rate of the pin placement is the highest when the pin length is between $2$ cm and $4$ cm. 

Note that $3$ cm is the scale of the main features for all objects in Figure~\ref{fig:model}, and thus one reasonable conjecture is that 
the pin placement has the highest success rate when the pin length is similar to the scale of the object's feature.
}

\begin{figure*}[!t]
\centering
\subfloat[$l$]{\includegraphics[width=0.45\linewidth]{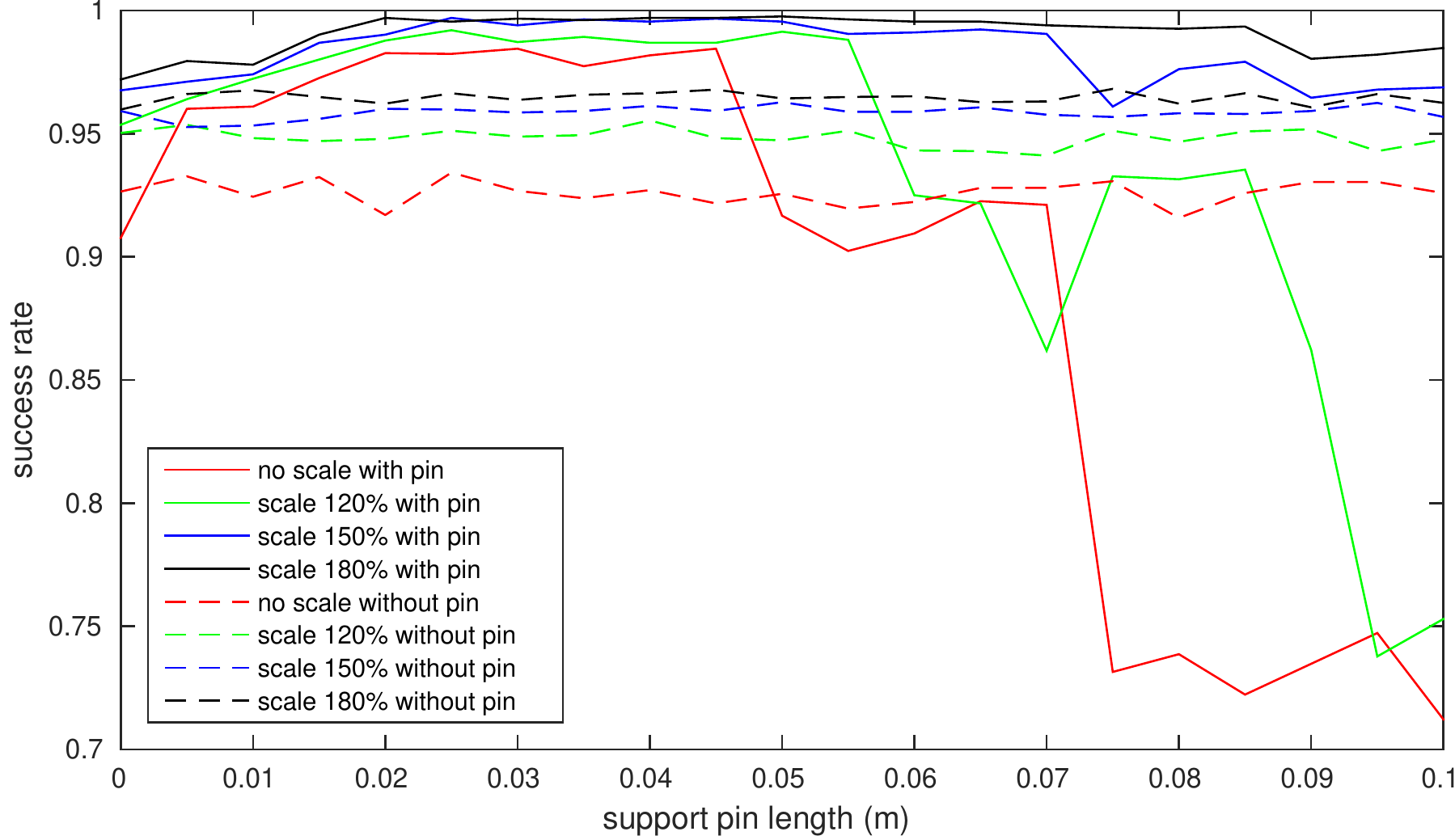}} \hfill
\subfloat[$el$]{\includegraphics[width=0.45\linewidth]{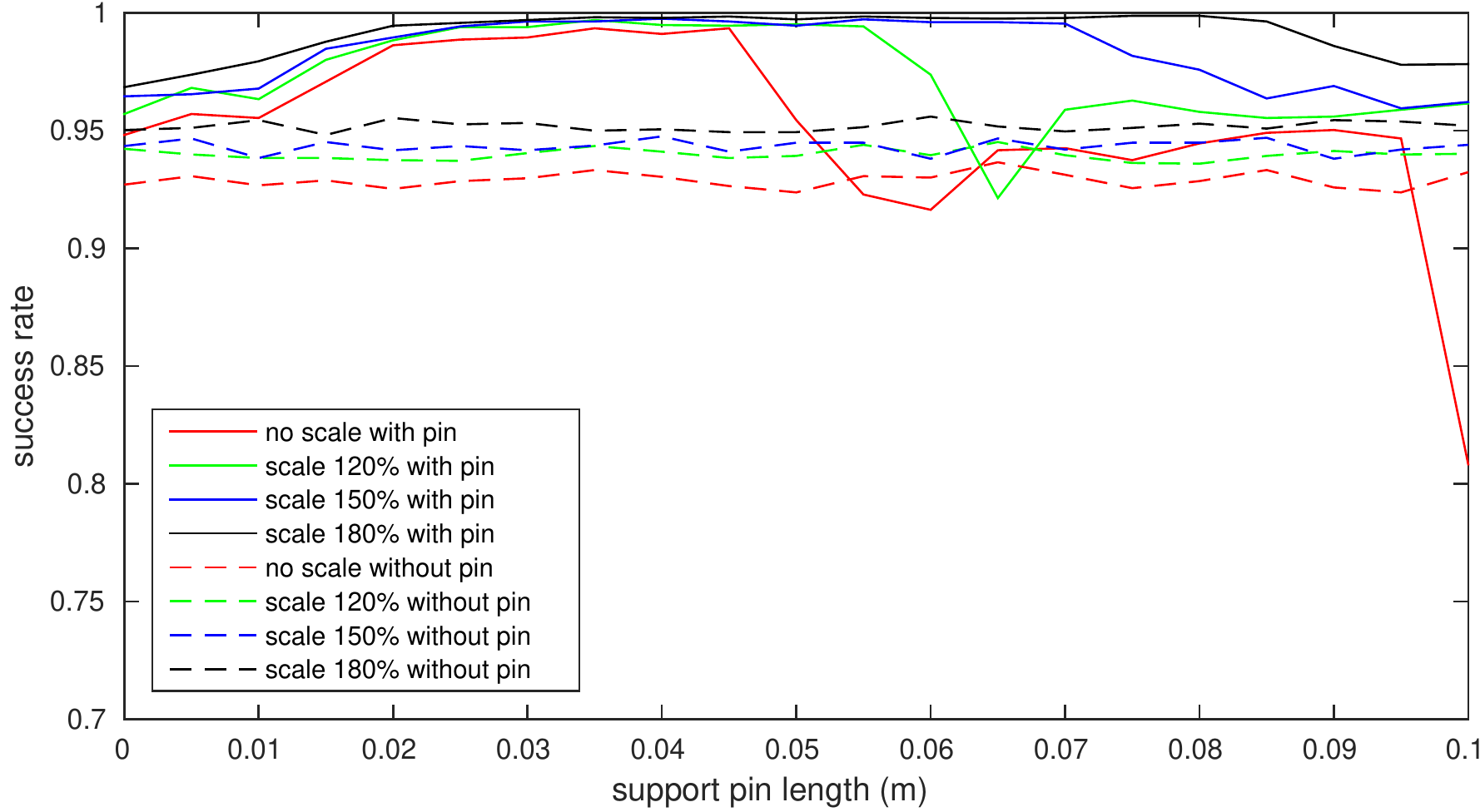}} \\
\subfloat[$3t$]{\includegraphics[width=0.45\linewidth]{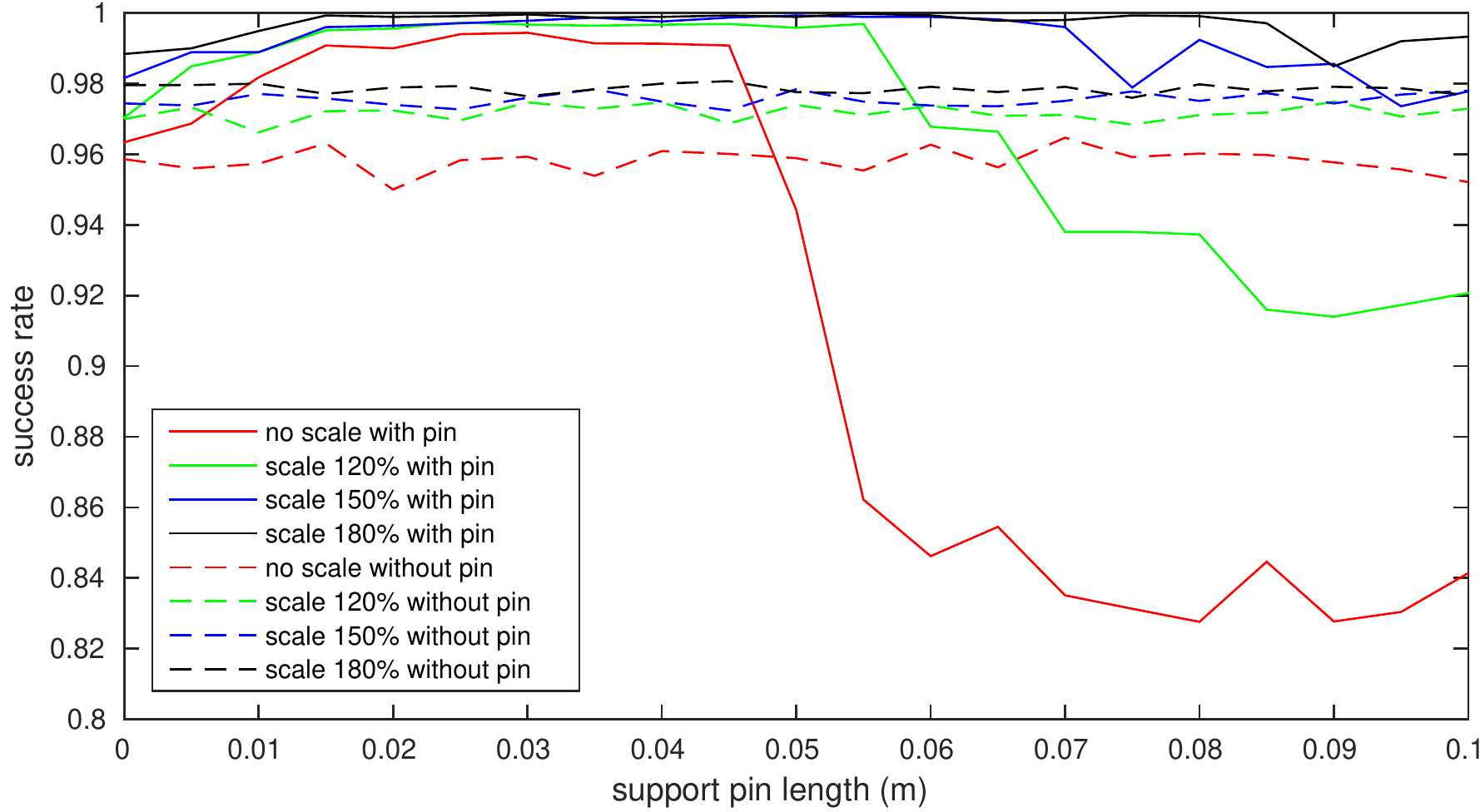}} \hfill
\subfloat[$3ts$]{\includegraphics[width=0.45\linewidth]{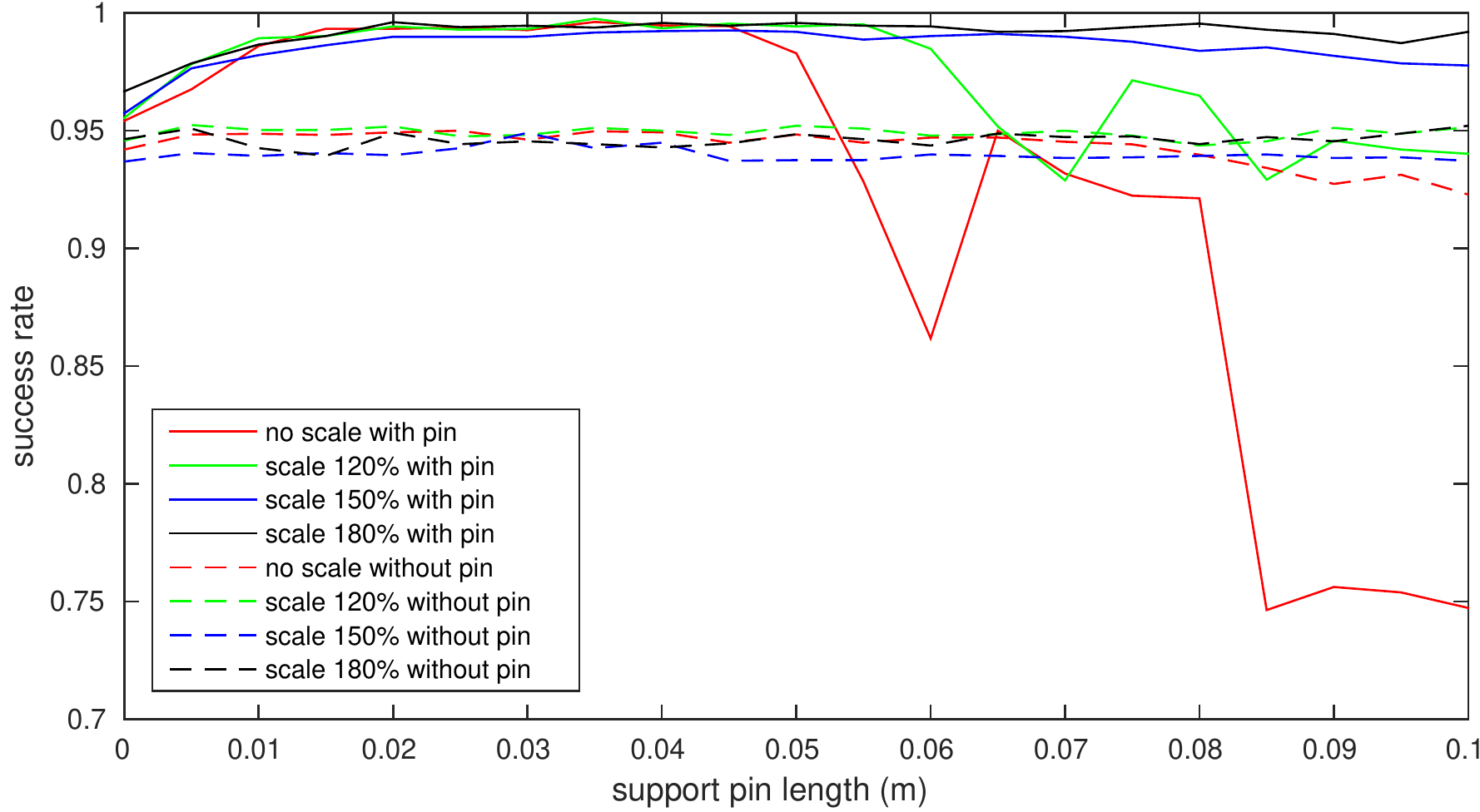}} \\
\subfloat[$3t2s$]{\includegraphics[width=0.45\linewidth]{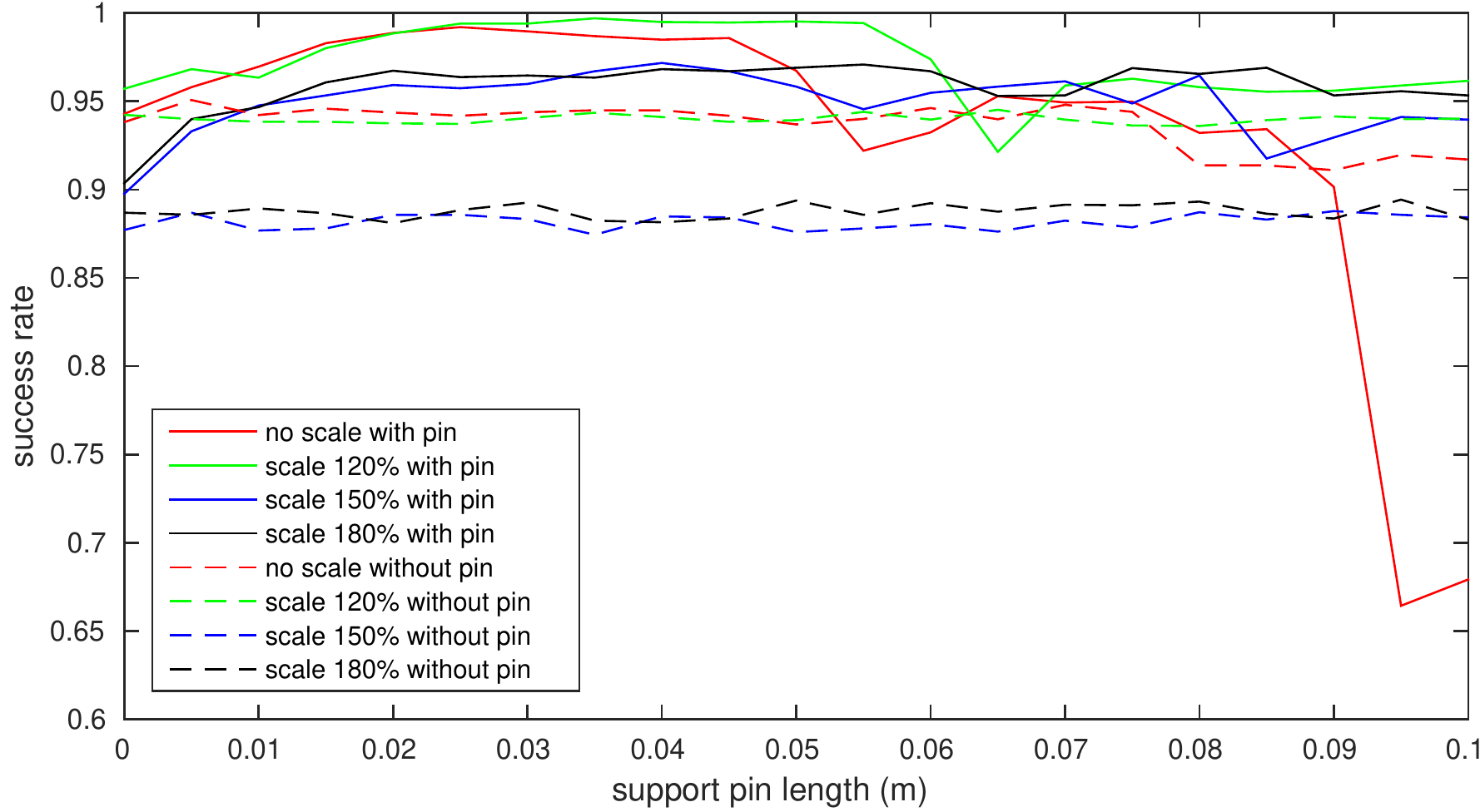}} \hfill
\subfloat[$cross$]{\includegraphics[width=0.45\linewidth]{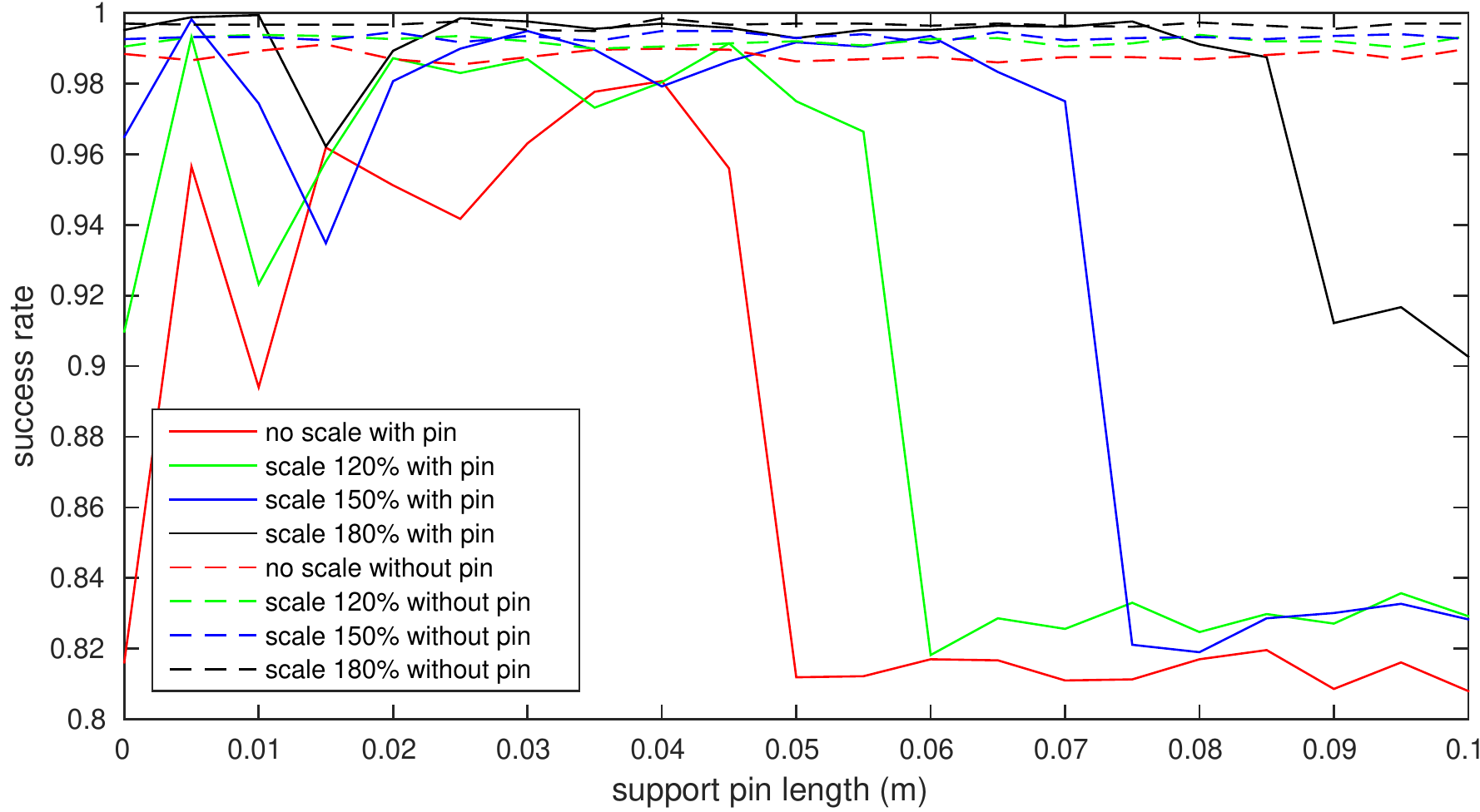}} \\
\subfloat[$t$]{\includegraphics[width=0.45\linewidth]{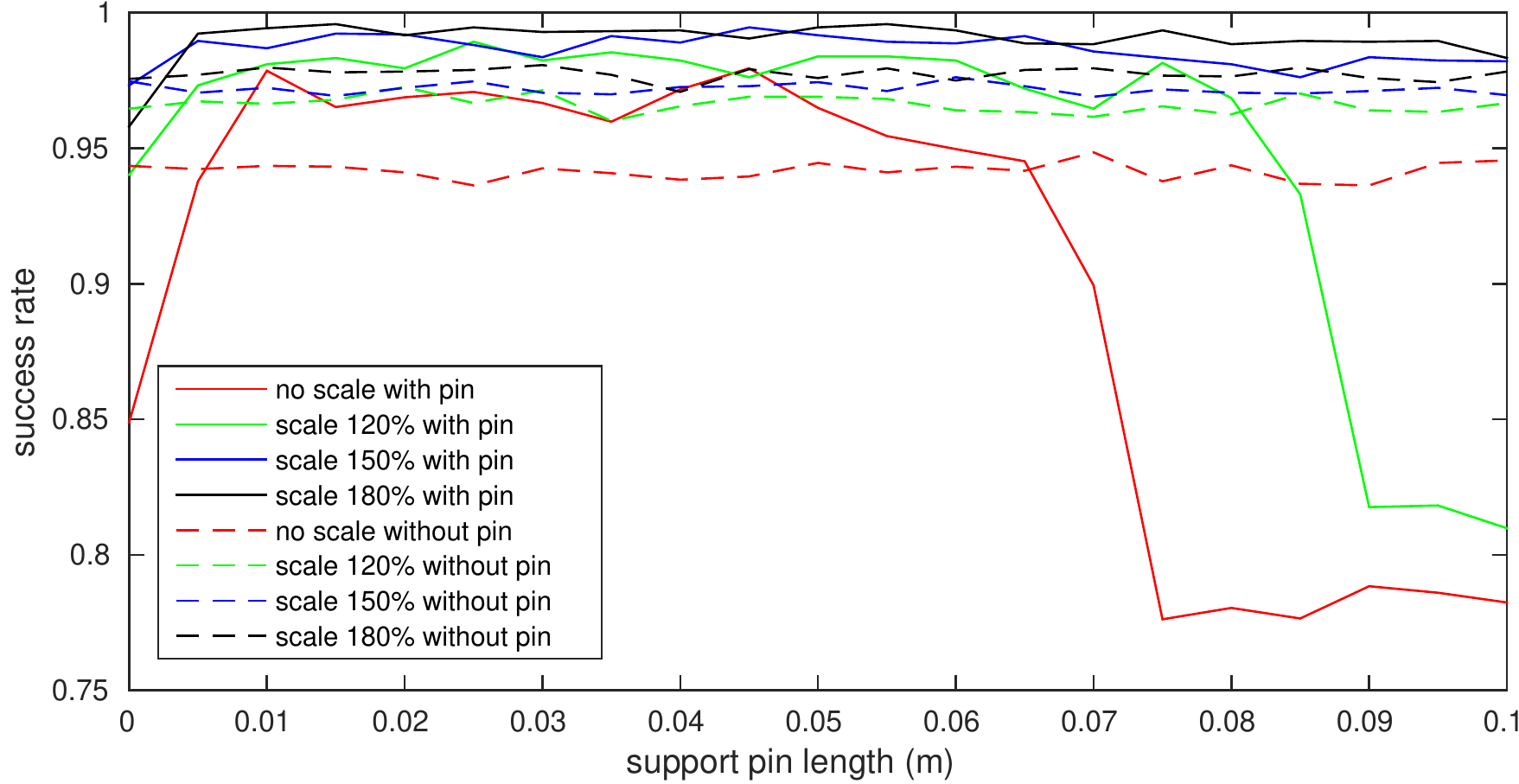}} \hfill
\subfloat[$1t1p$]{\includegraphics[width=0.45\linewidth]{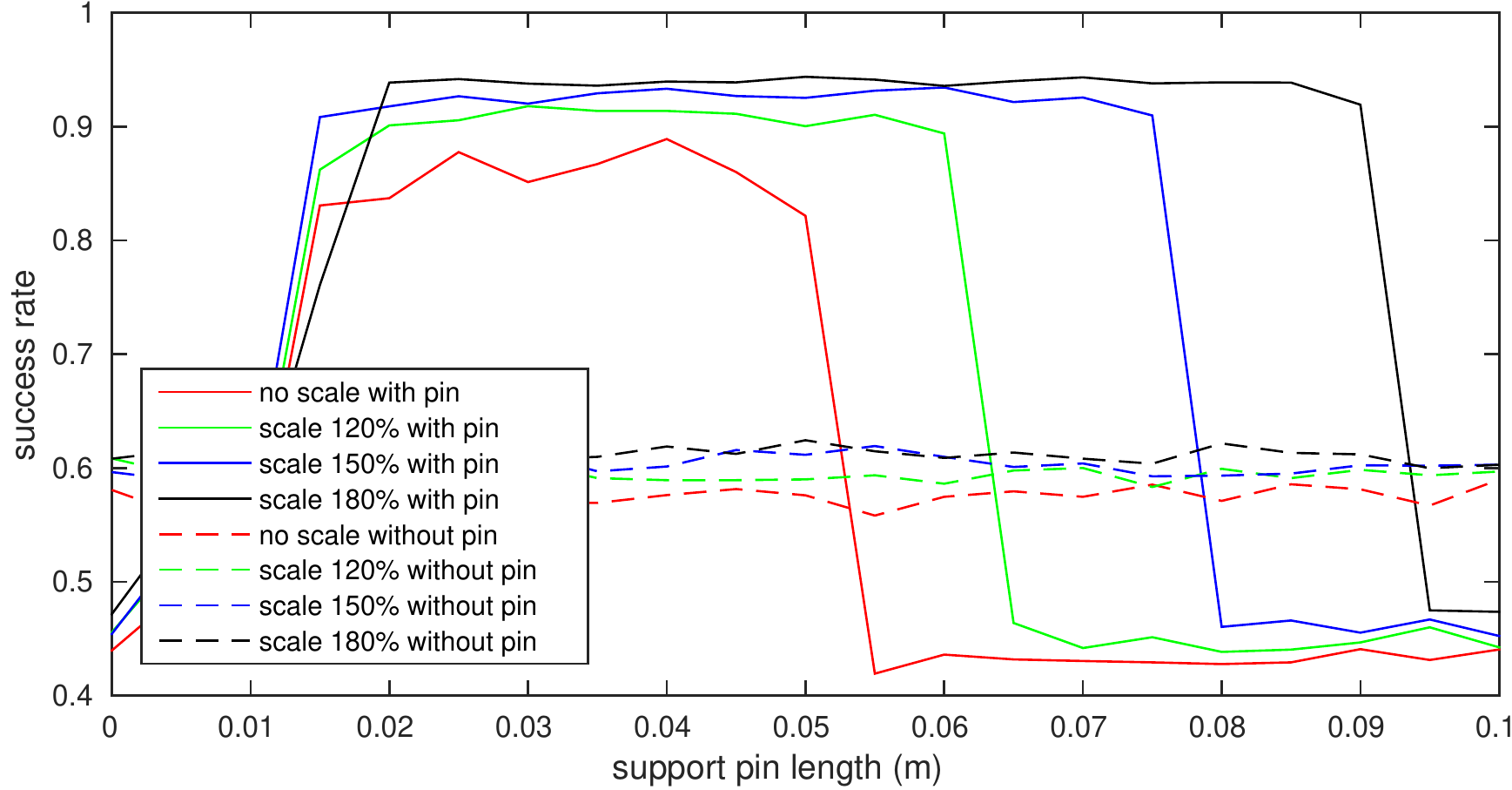}}
\caption{Relations between the success rate and the object's size while reorienting different models.}
\label{fig:objectSizeResults}
\end{figure*}

{\color{black}
\subsection{Relations between the Object Size and Success Rates}
We further investigate the relationship between the size of an object and the success rate of the orientation task. 
Given an object, we resize it from the center on four different scales ($100\%$, $120\%$, $150\%$ and $180\%$), and repeat the above experiments for each scale. The results are shown in Figure~\ref{fig:objectSizeResults}. For each scale of the objects, the curves illustrating the relation between the pin length and the success rate is of the similar shape as the curves shown in Figure~\ref{fig:stickLengthResults} for the unscaled objects. In addition, we find that when the object size increases, we can choose the pin length from a wider range but still achieve the optimal success rate. 
}

\subsection{Relations between the Grasp Density and Success Rates}
We also change the density of the total grasps by changing the number of
sampled directions as mentioned in Section~\ref{sec:algo:totalgrasp}. 

\begin{figure}[!t]
\centering
\subfloat[$l$]{\includegraphics[width=0.93\linewidth]{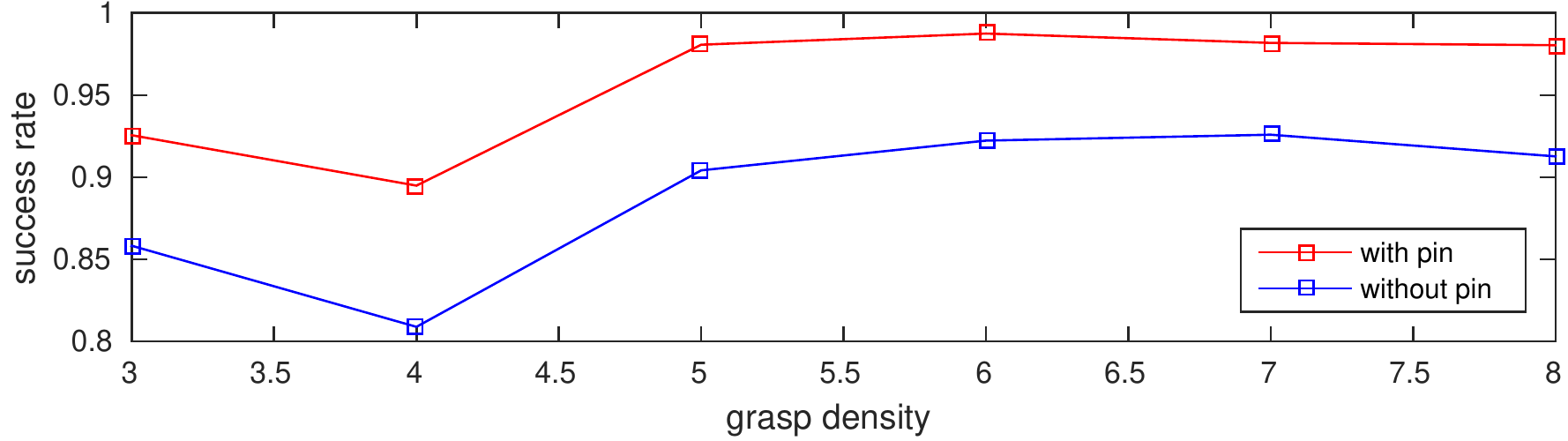}} \vspace*{-0.08in} \\
\subfloat[$el$]{\includegraphics[width=0.93\linewidth]{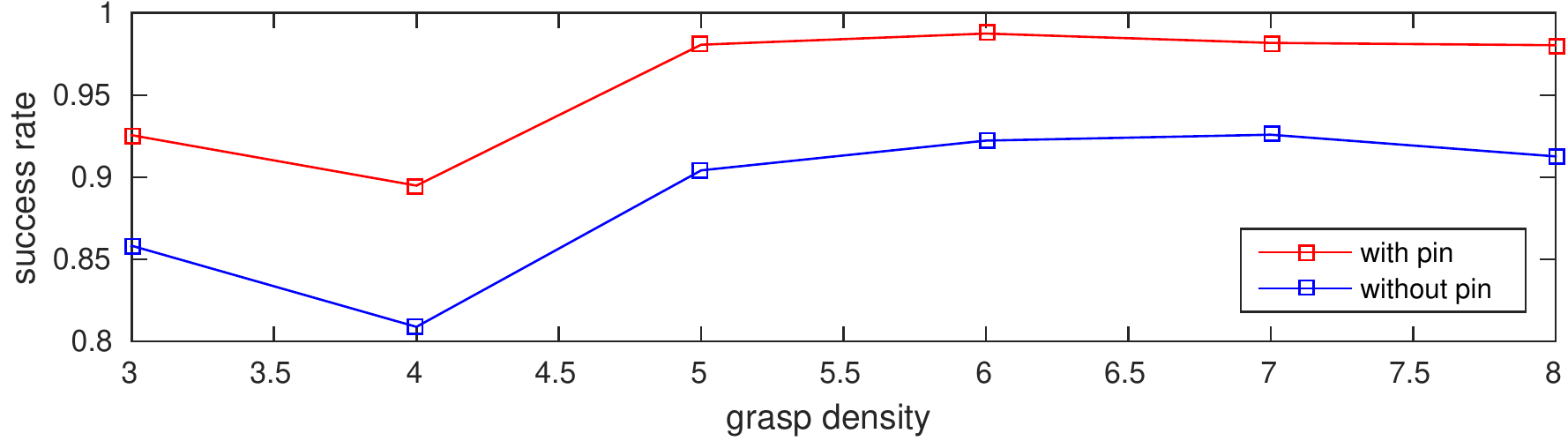}} \vspace*{-0.08in} \\
\subfloat[$3t$]{\includegraphics[width=0.93\linewidth]{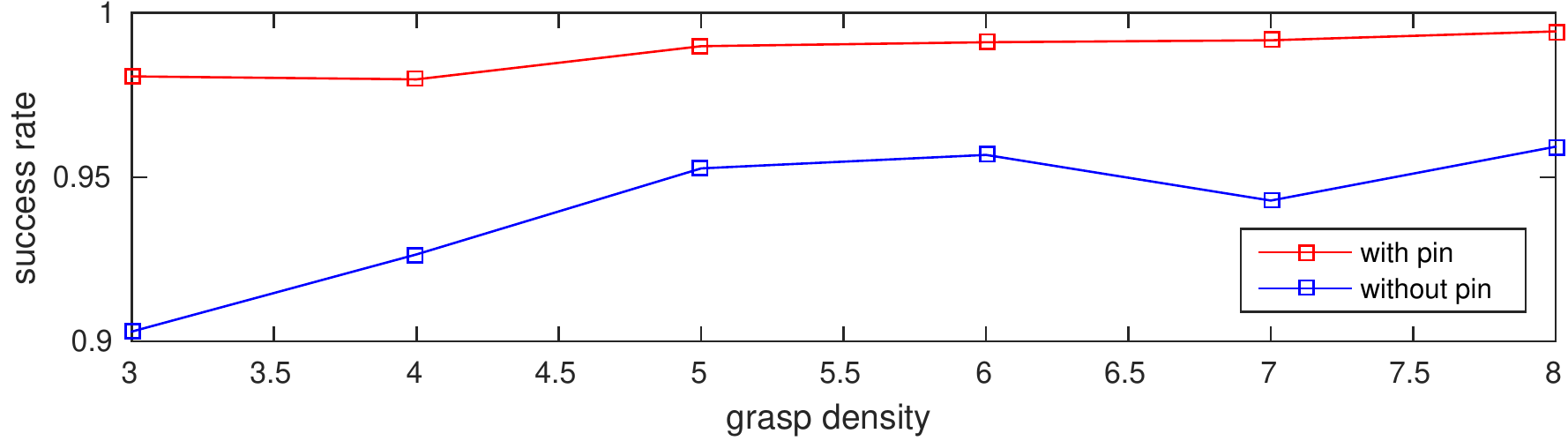}} \vspace*{-0.08in} \\
\subfloat[$3ts$]{\includegraphics[width=0.93\linewidth]{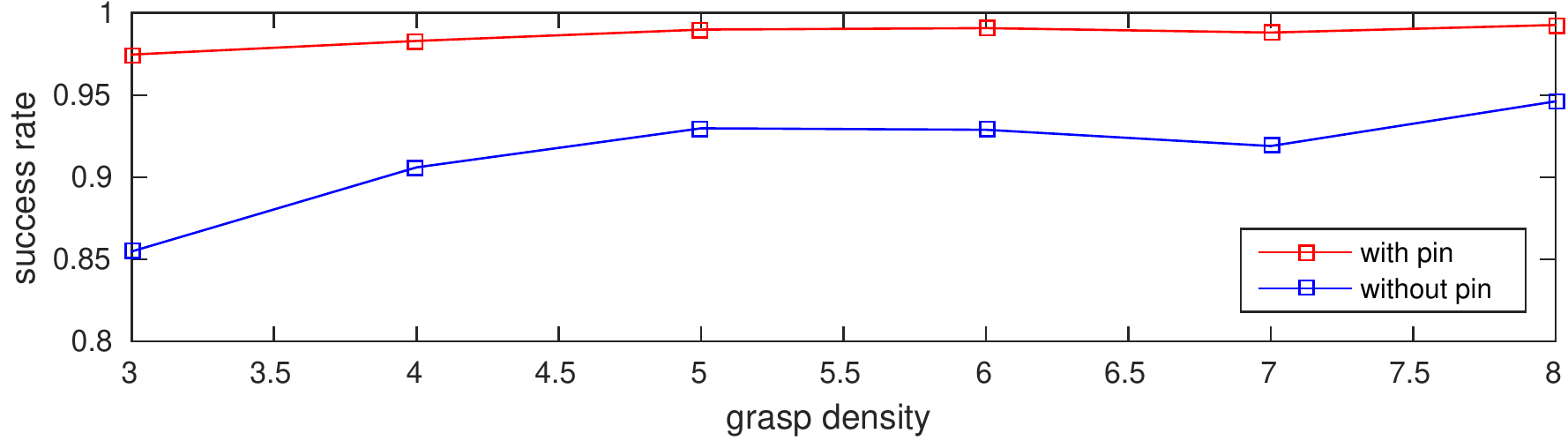}} \vspace*{-0.08in} \\
\subfloat[$3t2s$]{\includegraphics[width=0.93\linewidth]{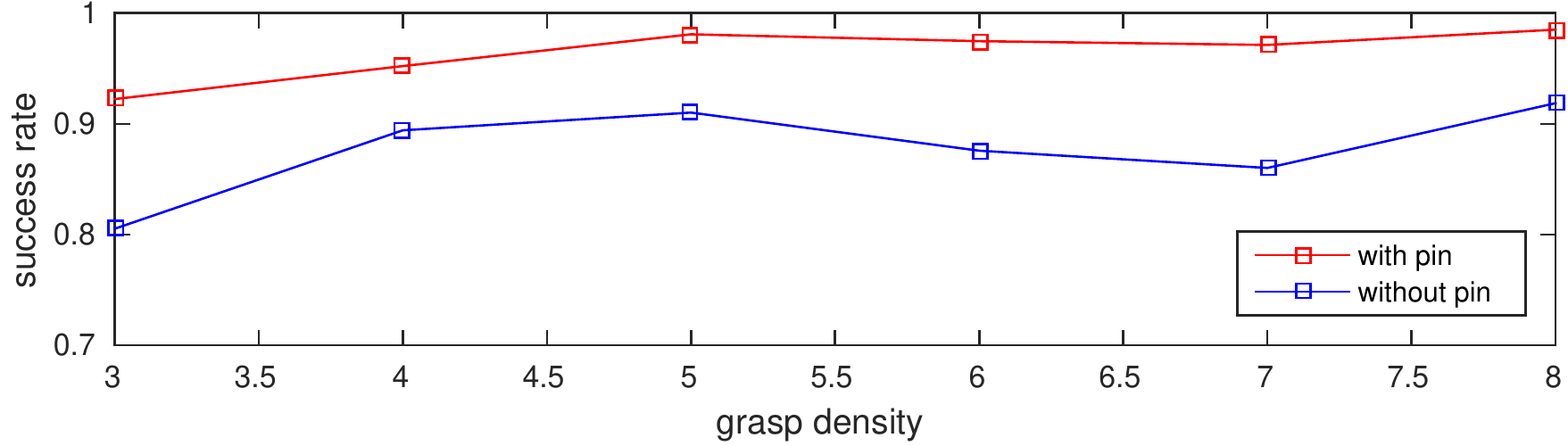}} \vspace*{-0.08in} \\
\subfloat[$cross$]{\includegraphics[width=0.93\linewidth]{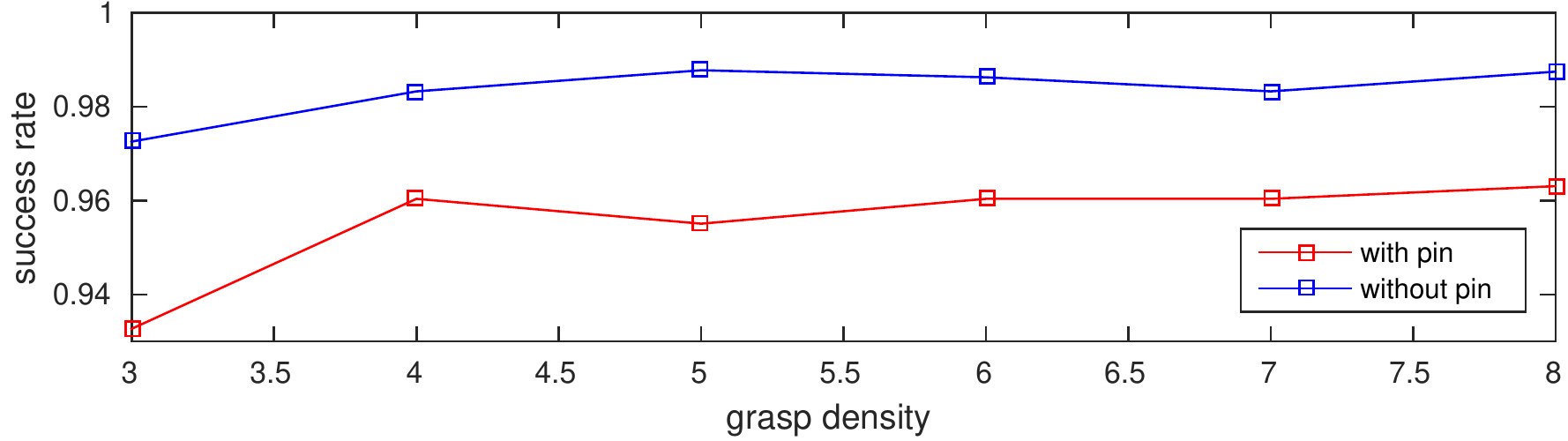}} \vspace*{-0.08in} \\
\subfloat[$t$]{\includegraphics[width=0.93\linewidth]{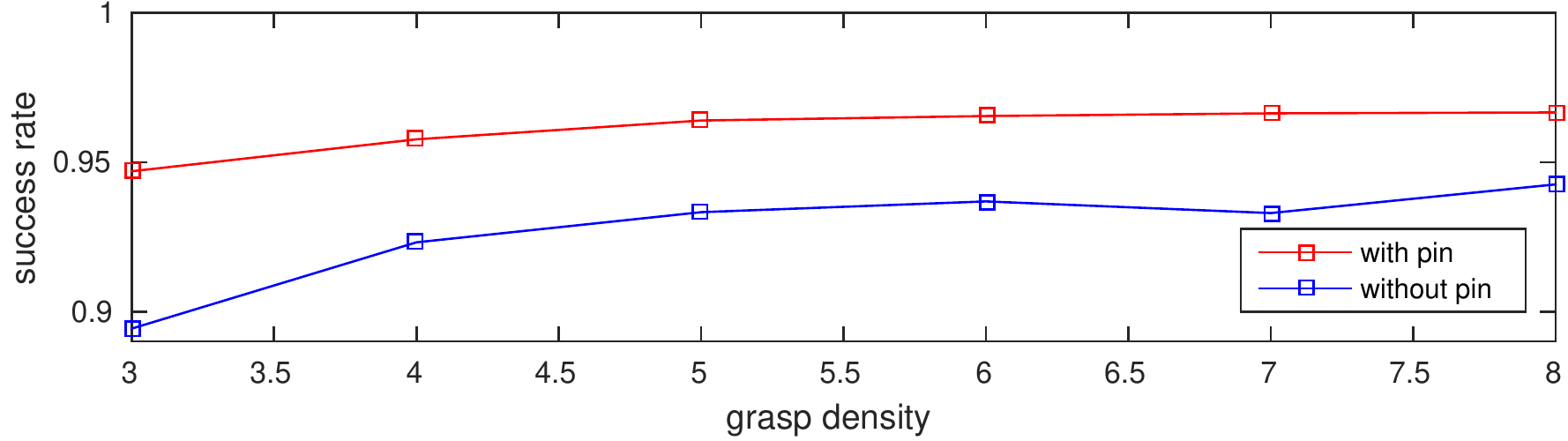}} \vspace*{-0.08in} \\
\subfloat[$1t1p$]{\includegraphics[width=0.93\linewidth]{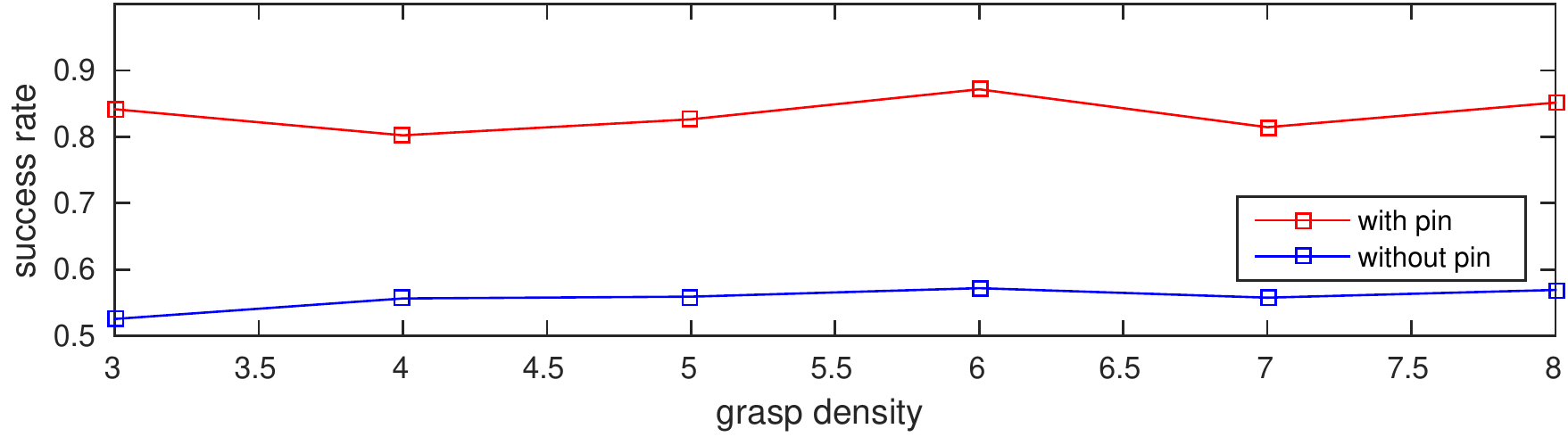}}
\caption{Relations between the success rate and the changing grasp density while
reorienting the models.}
\label{fig:densityResult}
\end{figure}

\begin{figure}[!t]
\centering
\subfloat[$l$]{\includegraphics[width=0.93\linewidth]{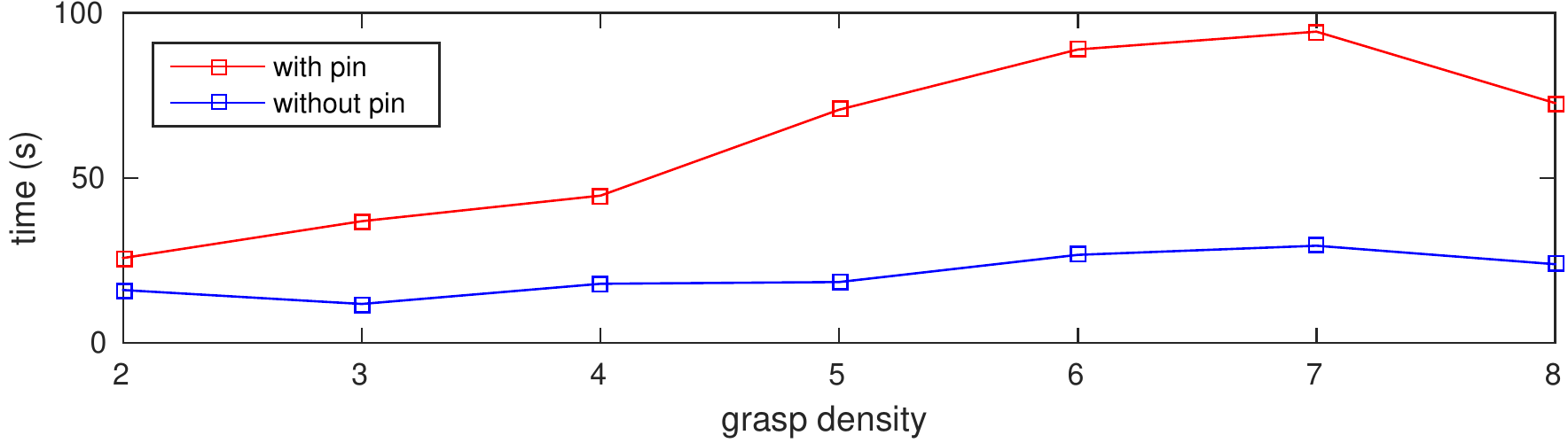}} \vspace*{-0.08in} \\
\subfloat[$el$]{\includegraphics[width=0.93\linewidth]{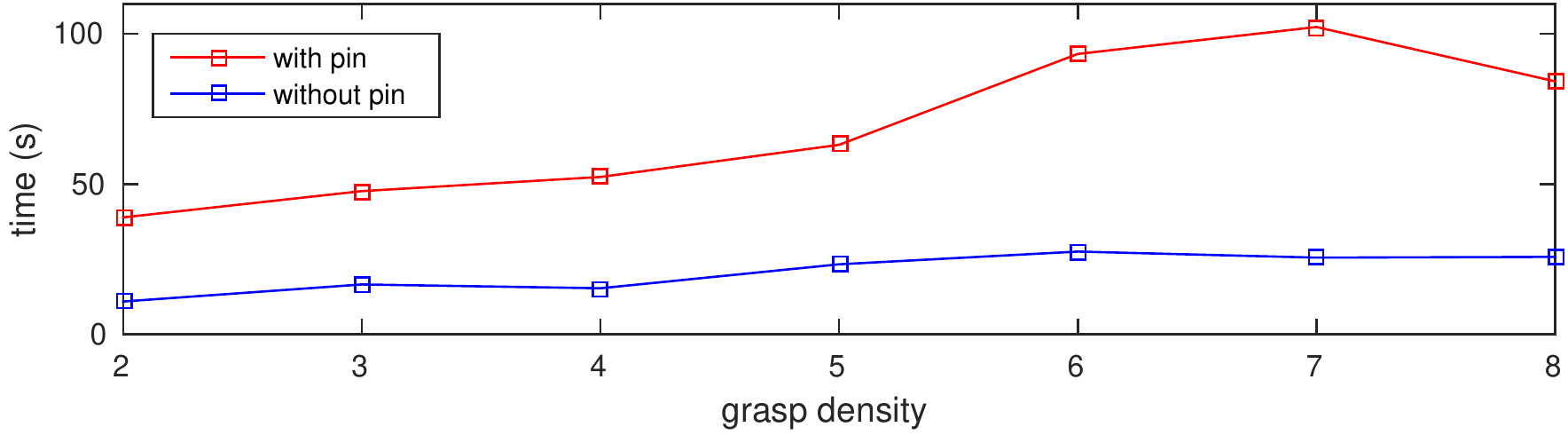}} \vspace*{-0.08in}\\
\subfloat[$3t$]{\includegraphics[width=0.93\linewidth]{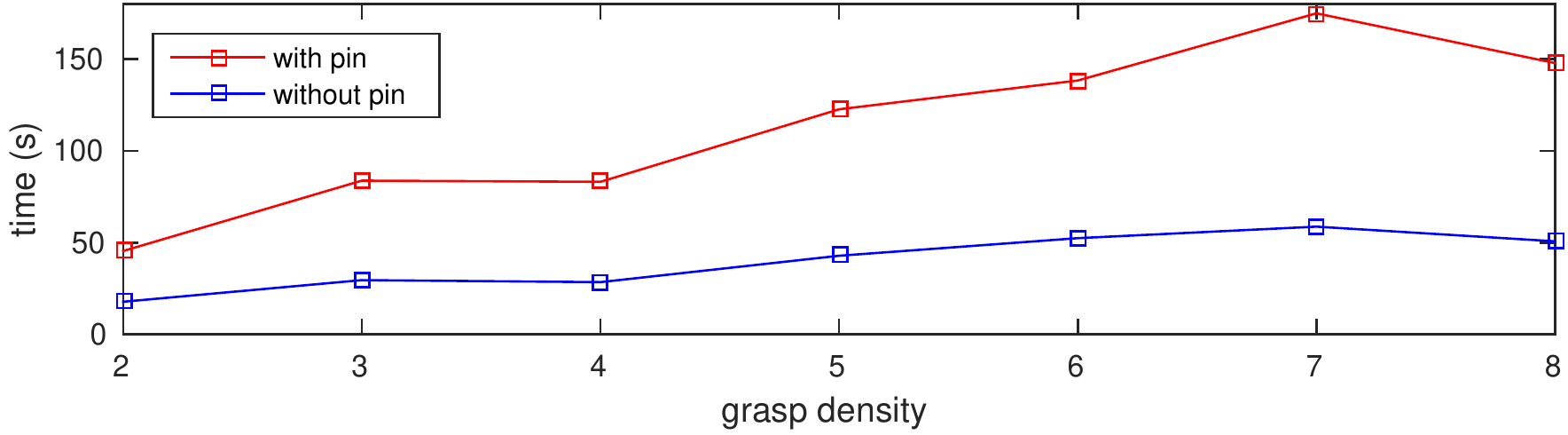}}  \vspace*{-0.08in} \\
\subfloat[$3ts$]{\includegraphics[width=0.93\linewidth]{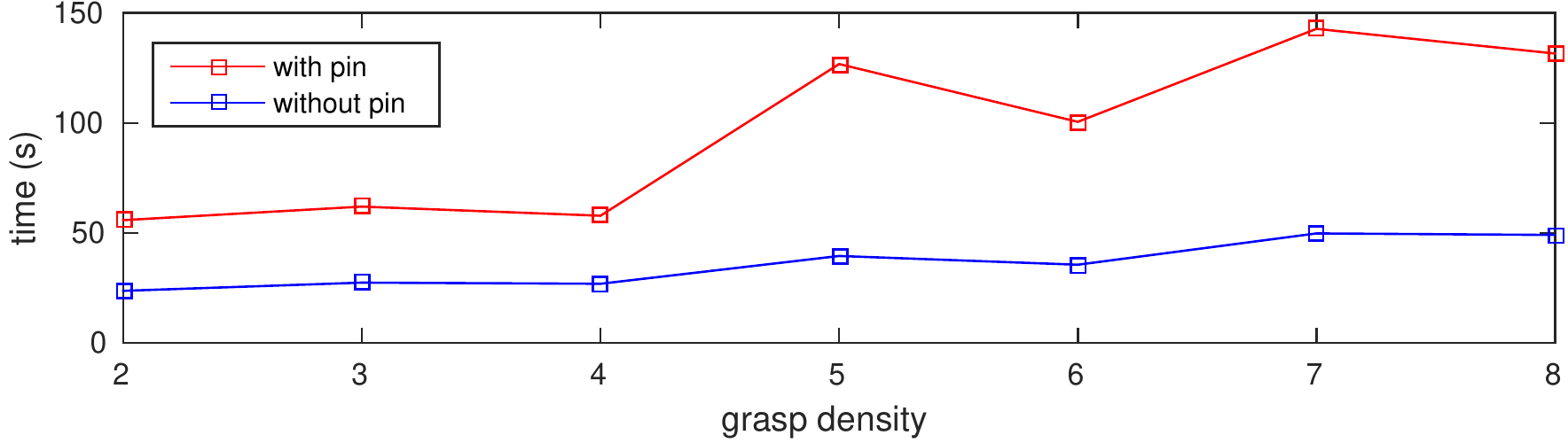}} \vspace*{-0.08in} \\
\subfloat[$3t2s$]{\includegraphics[width=0.93\linewidth]{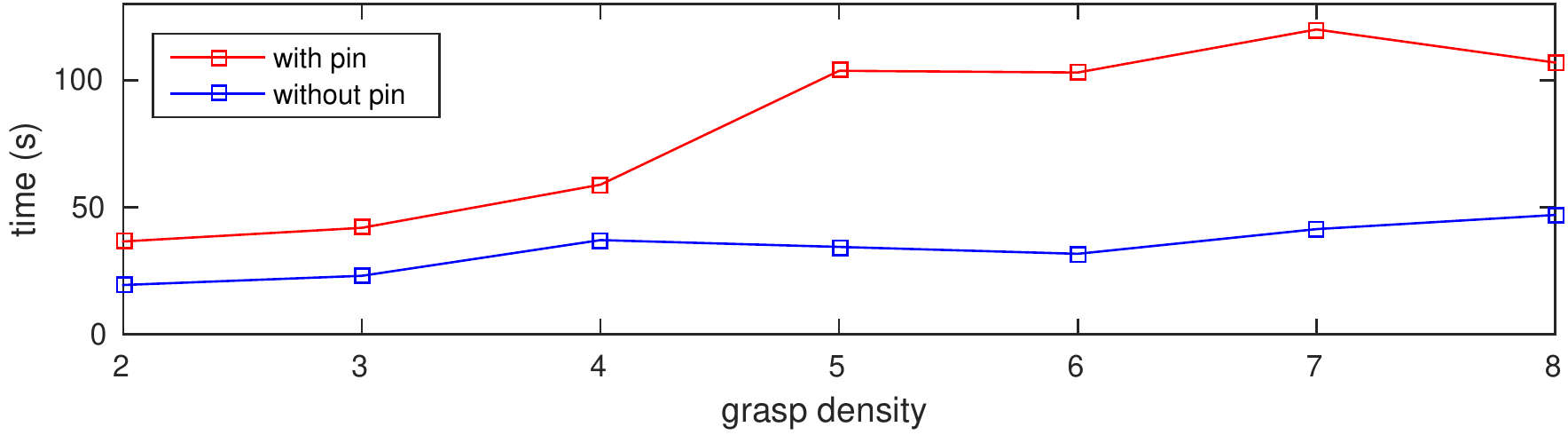}} \vspace*{-0.08in} \\
\subfloat[$cross$]{\includegraphics[width=0.93\linewidth]{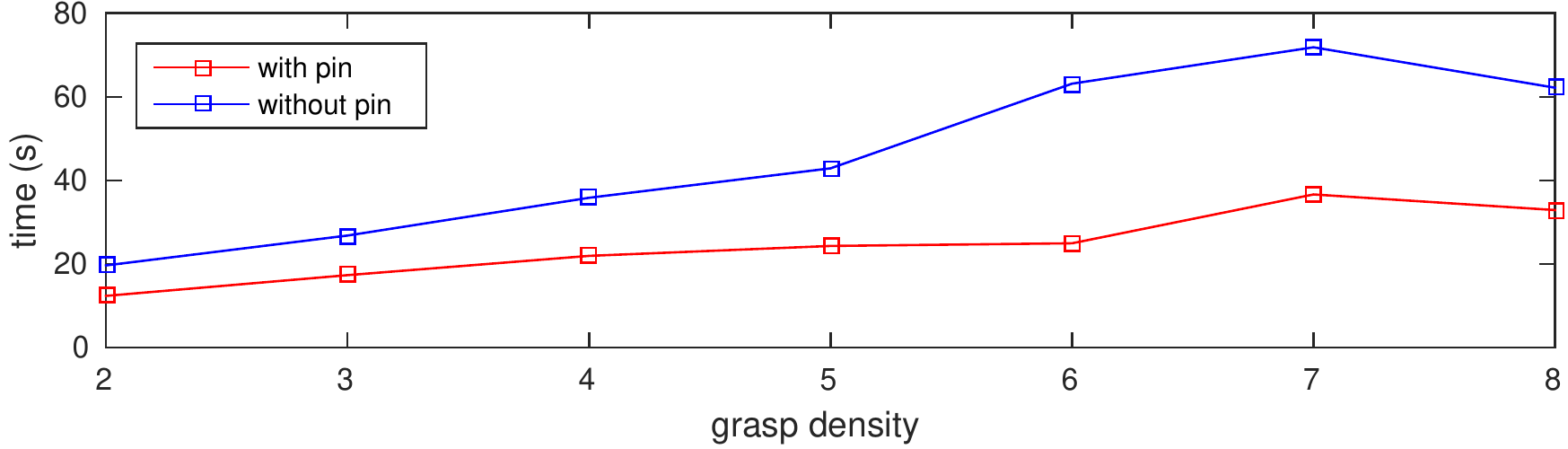}} \vspace*{-0.08in} \\
\subfloat[$t$]{\includegraphics[width=0.93\linewidth]{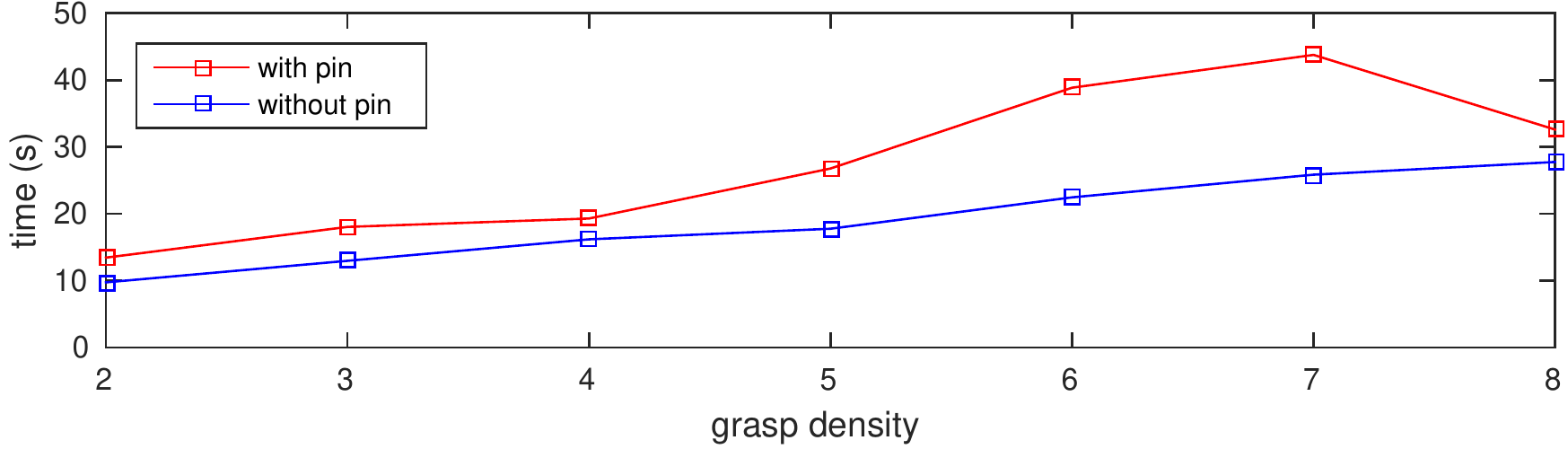}} \vspace*{-0.08in} \\
\subfloat[$1t1p$]{\includegraphics[width=0.93\linewidth]{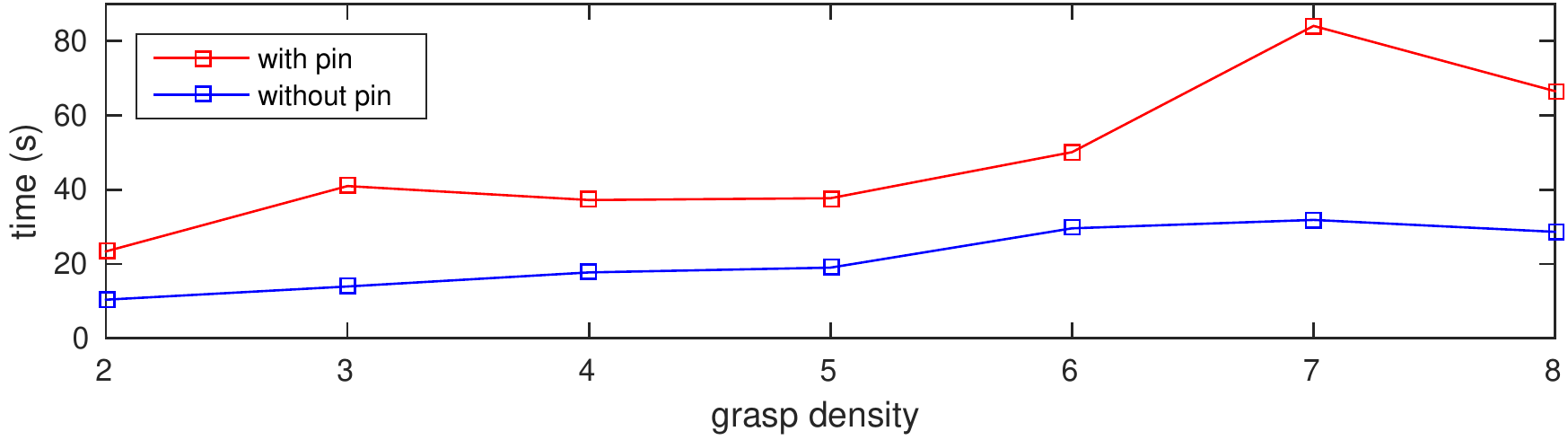}}
\caption{Relations between the graph build time and the changing grasp density while
reorienting the models.}
\label{fig:densityBuildGraphTimeResult}
\end{figure}

\begin{figure}[!t]
\centering
\subfloat[$l$]{\includegraphics[width=0.93\linewidth]{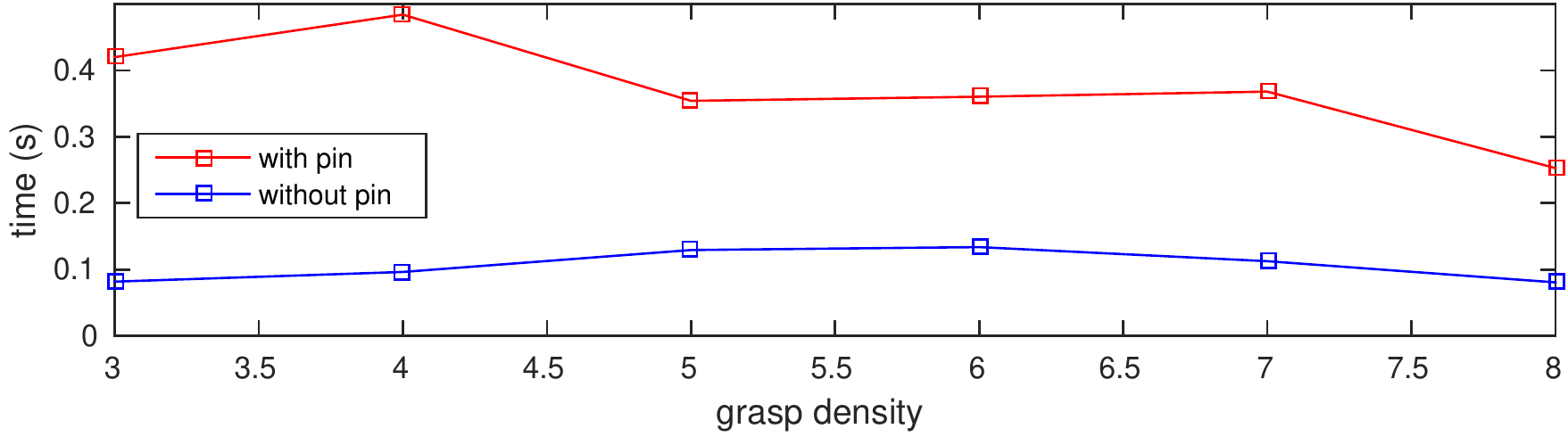}}  \vspace*{-0.08in} \\
\subfloat[$el$]{\includegraphics[width=0.93\linewidth]{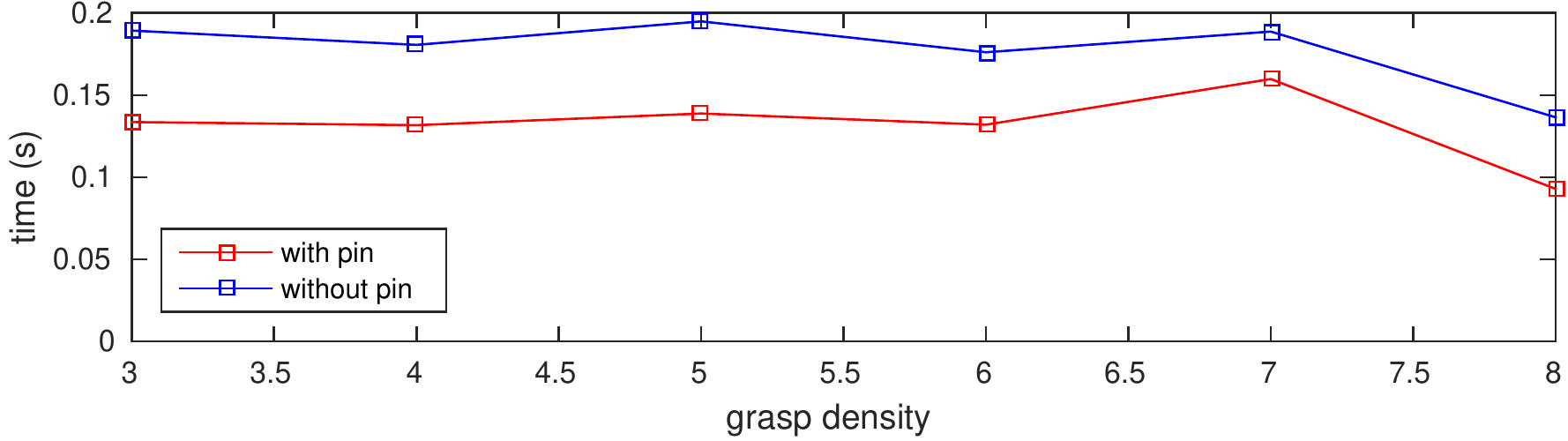}} \vspace*{-0.08in} \\
\subfloat[$3t$]{\includegraphics[width=0.93\linewidth]{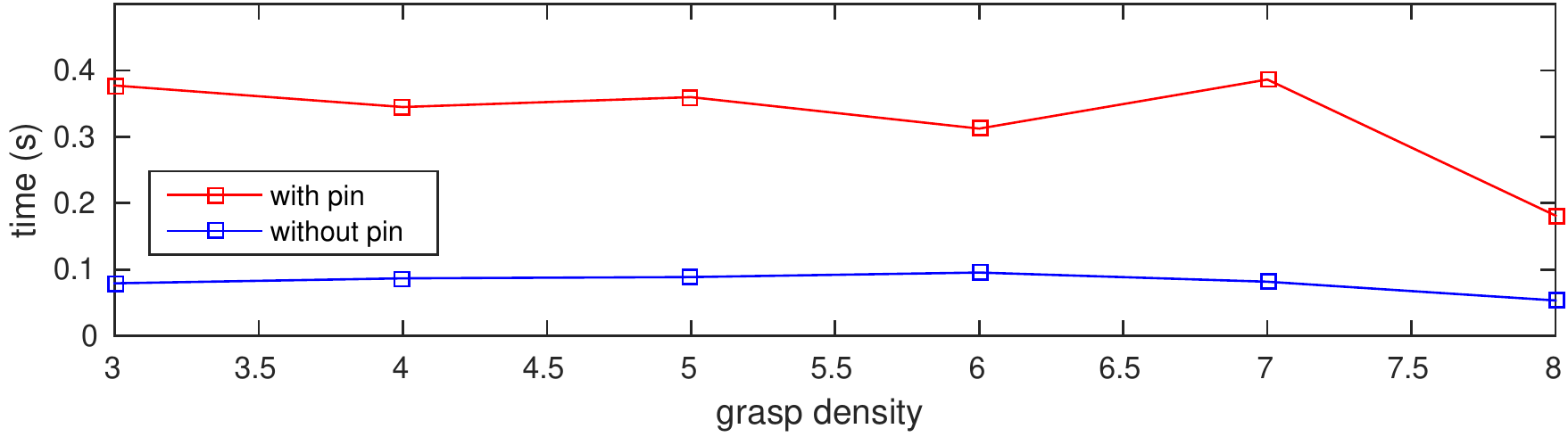}} \vspace*{-0.08in} \\
\subfloat[$3ts$]{\includegraphics[width=0.93\linewidth]{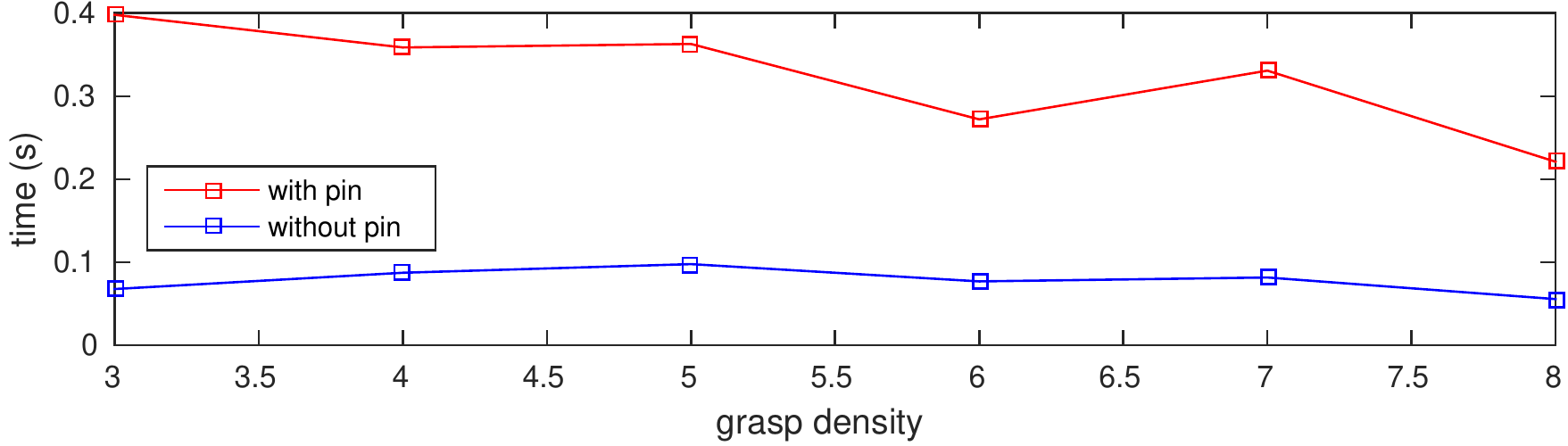}} \vspace*{-0.08in} \\
\subfloat[$3t2s$]{\includegraphics[width=0.93\linewidth]{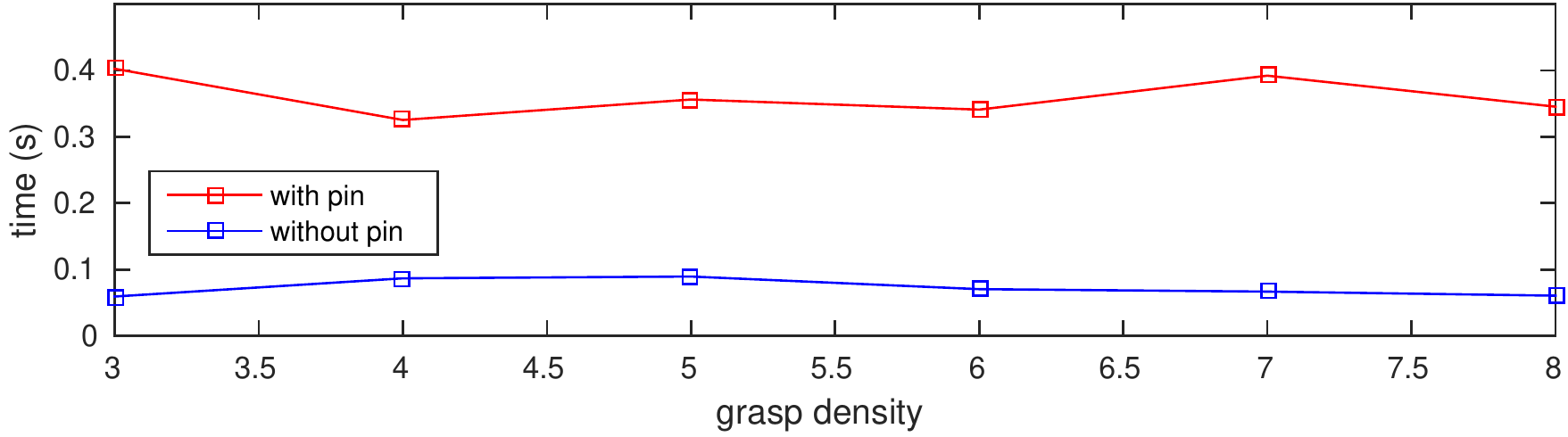}} \vspace*{-0.08in} \\
\subfloat[$cross$]{\includegraphics[width=0.93\linewidth]{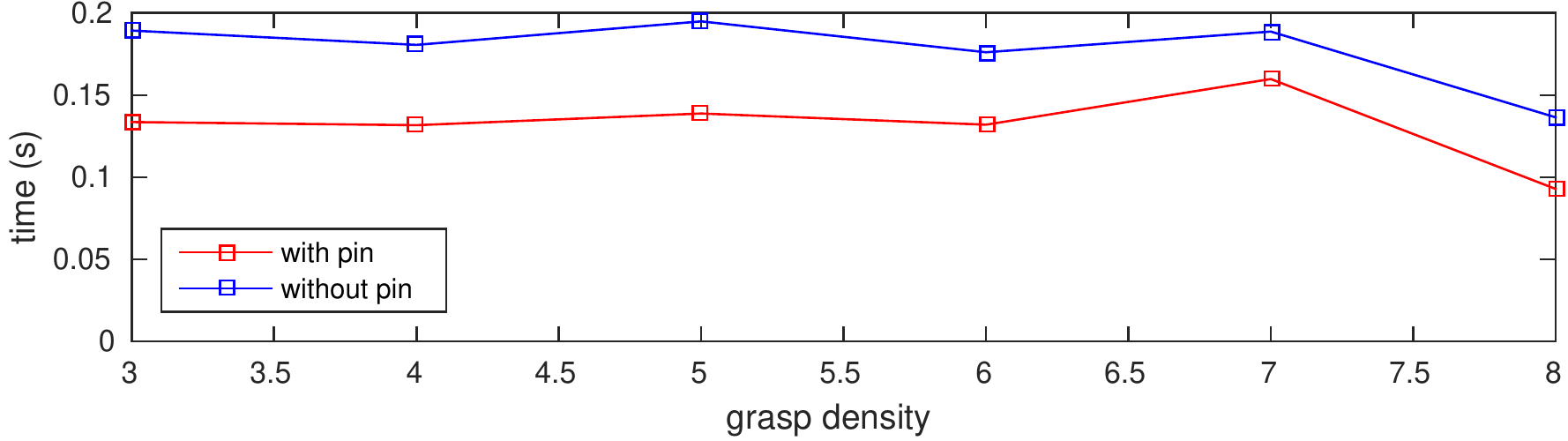}}  \vspace*{-0.08in} \\
\subfloat[$t$]{\includegraphics[width=0.93\linewidth]{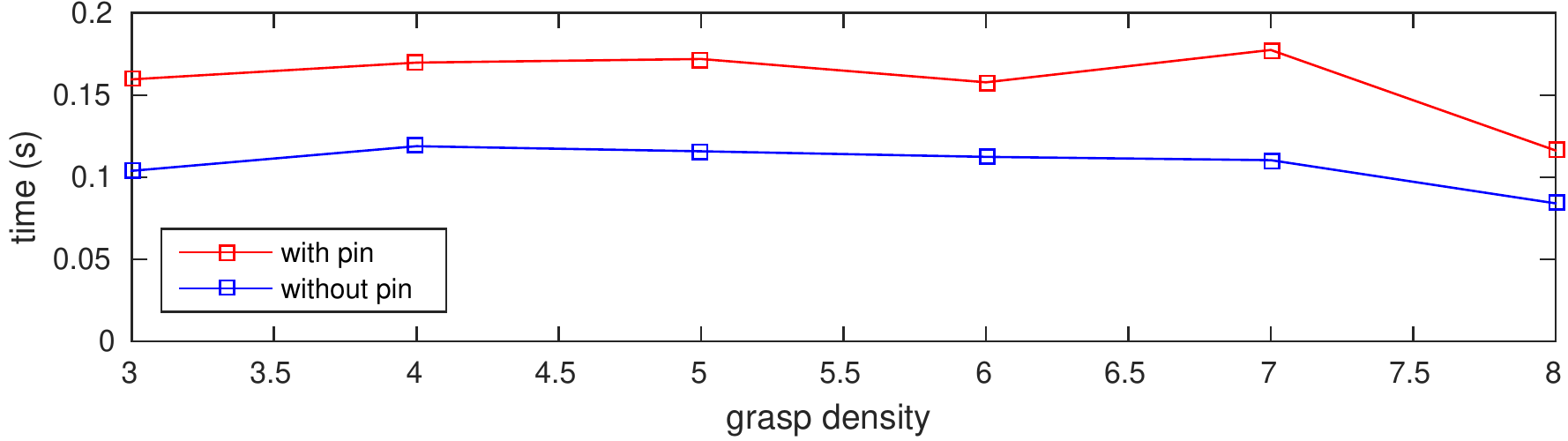}} \vspace*{-0.08in} \\
\subfloat[$1t1p$]{\includegraphics[width=0.93\linewidth]{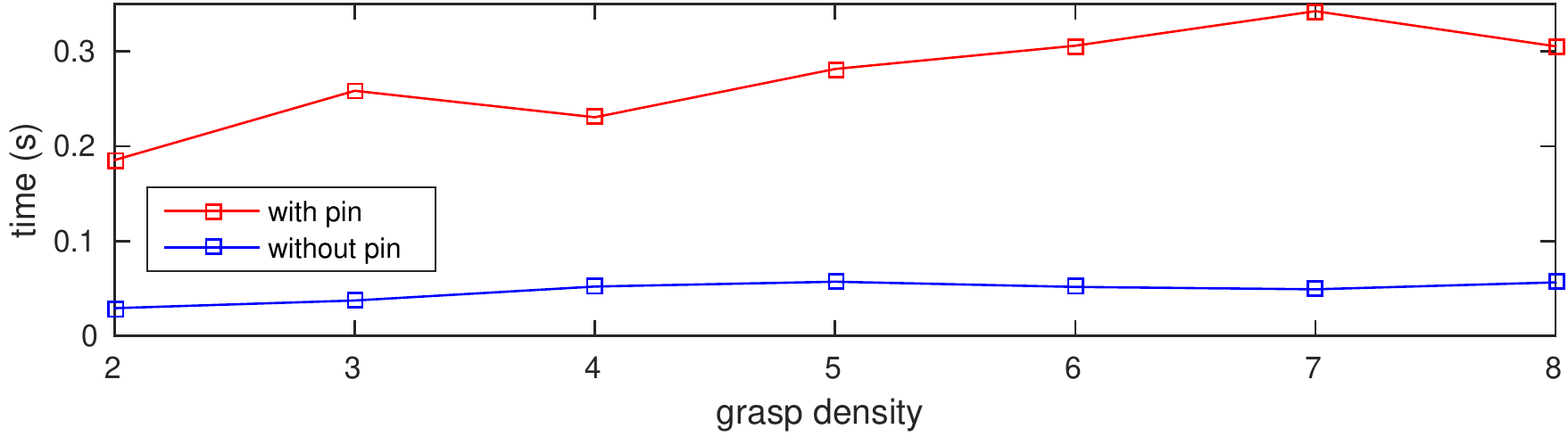}}
\caption{Relations between the graph search time and the changing grasp density while
reorienting the models.}
\label{fig:densitySearchGraphTimeResult}
\end{figure}

{\color{black}
The results in Figure~\ref{fig:densityResult} illustrate the changing success rate with
respect to various grasp density. In general the success rate of the orientation
task improves as the number of sampled directions increases from $3$ to $8$, and
the small fluctuation is due to the randomness in the experiment. We observe that the 
reorientation success rate increases when the grasp density increases. However, the performance 
improvement is small for grasp densities larger than $6$. 

We also illustrate the grasp density's influence on the timing cost of the regrasp graph construction and the regrasp grasp online search in Figure~\ref{fig:densityBuildGraphTimeResult} and Figure~\ref{fig:densitySearchGraphTimeResult}, respectively. We can observe that the time cost while using the pin placement is generally slower than the time cost while using the planar placement, because the regrasp graph for the pin placement is more complicated. The $cross$ model is the only exception, where the regrasp graph of the pin placement has fewer connectivity due to the reason
discussed in Section~\ref{sec:experiment:avgsuccess}, and thus the timing cost of the pin placement is also lower.

According to these results, we conclude that a grasp density of $6$ can make a good tradeoff between the reorientation success rate and the time cost. 
}

\subsection{Comparison of Regrasp Sequence Lengths}
\label{sec:experiment:avglength}
\begin{figure}[!t]
\centering
\includegraphics[width=1\linewidth]{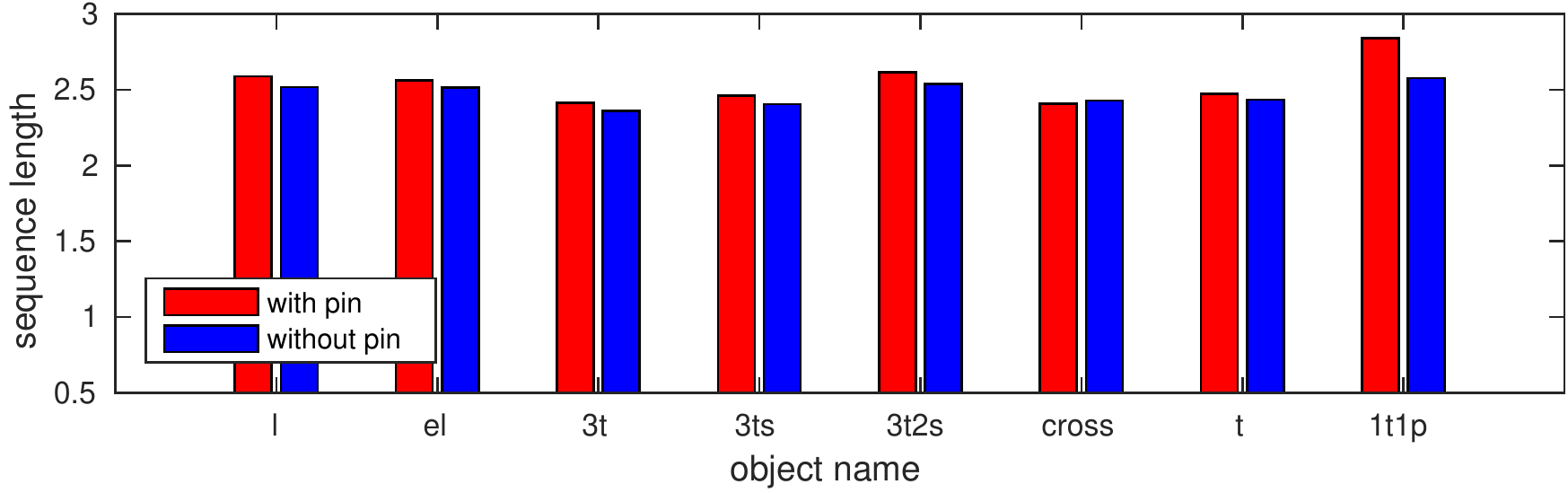}
\caption{Comparison between the average sequence length of the orientation task,
while using two different placement settings. The red bars are the results
using the added pin for placement, while the blue bars are the results only
using the planar surface for support.}
\label{fig:sequencelength}
\end{figure}

We also compare the number of regrasps used while using the pin placement to
the number of regrasps used while using the planar placement. As shown in Figure~\ref{fig:sequencelength}, we can
see that in general the two different placement settings will result in regrasp
sequences with similar lengths. For the pot lid shape object $1t1p$, the
sequence length while using the pin placement is significantly longer than the sequence length while
using the planar placement. This is because the $1t1p$ is difficult to be
oriented, and thus for the planar placement setting, many reorientation trials
fail and are not counted in the computation of average length of regrasp
sequences. While using the pin placement, the reorientation task has a higher
success rate for difficult trials, but may require more regrasps to achieve the
reorientation in such difficult cases.

\subsection{Spatial Distribution of Success Rates and Regrasp Sequence Lengths}
Besides the average success rate and average regrasp sequence length computed
over all trials as demonstrated in Section~\ref{sec:experiment:avgsuccess} and
Section~\ref{sec:experiment:avglength}, we also compute the average success rate
and regrasp sequence length over the 10 trials for each grid in the working
area. These data describe the spatial distribution of the success rate and
sequence length for the reorientation task, and help us to have a better
understanding about the difference between the different placement settings.

\begin{figure}[!t]
\centering
\includegraphics[width=1\linewidth]{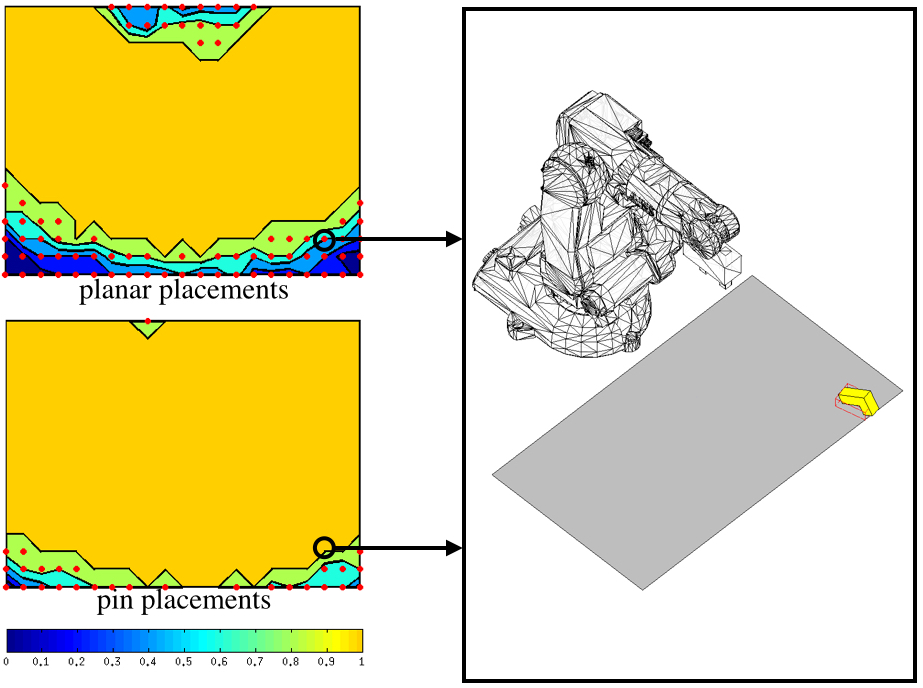}
\caption{Comparison of the spatial distribution of the success rates between the
pin placement and the planar placement while orienting the $l$ model. The robot
is standing in the north side of the rectangle working space. The red points are
the grid points where the reorientation task may fail, i.e., success rate is
less than 100\%. Different colors are used to visualize different values of
success rate, ranging from 0 to 1. The right sub-figure shows one instance of
the orientation task that fails while using the planar placement but succeeds
while using the pin placement.}
\label{fig:spatialsuccess}
\end{figure}

\begin{figure}[!t]
\centering
\subfloat[$1t1p$ planar placements]{\includegraphics[width=0.48\linewidth]{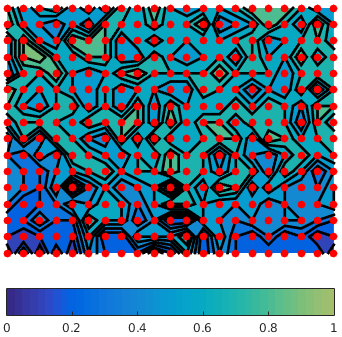}}
\subfloat[$1t1p$ pin placements]{\includegraphics[width=0.48\linewidth]{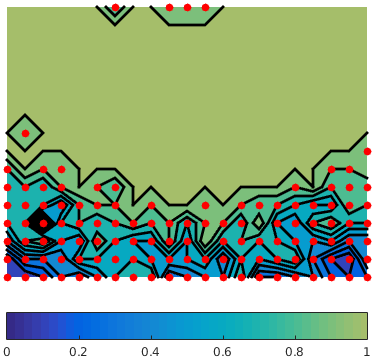}} \\
\subfloat[reachability]{\includegraphics[width=0.6\linewidth]{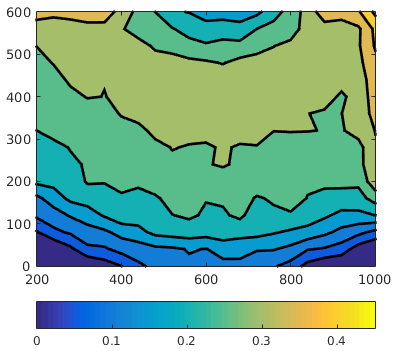}}
\caption{Comparison of the spatial distribution of the success rates between the pin placement and the planar placement while orienting the pot lid like object $1t1p$. The red points are
the grid points where the reorientation task may fail, i.e., success rate is
less than 100\%. Different colors are used to visualize different values of
success rate, ranging from 0 to 1. We also compute the reachability of the single arm, in order to demonstrate the difference between the spatial distribution of the success rates and the robotic arm's reachability.
}
\label{fig:spatialsuccessMore}
\end{figure}

{\color{black}
First, we compare the spatial distributions of the success rate while performing the reorientation task between two different placement settings for two objects $l$ and $1t1p$, and the results are shown in Figure~\ref{fig:spatialsuccess} and Figure~\ref{fig:spatialsuccessMore}. In Figure~\ref{fig:spatialsuccess},
we also illustrate one trial where the reorientation task fails in the planar
placement setting but succeeds in the pin placement setting. 
From the result for $l$ model in Figure~\ref{fig:spatialsuccess}, we can observe that, while using the pin placement, the robot will have a larger region in the
workspace where the success rate is 100\%. For other regions, the pin placement
also provides a higher success rate than that of the planar placement. 
The contrast becomes more obvious for the pot lid like object $1t1p$ in Figure~\ref{fig:spatialsuccessMore}. 
We observe that the planar placement always has a non-zero failure possibility over the entire workspace, while the pin placement has a $100\%$ success rate 
over a large part of the workspace.

Note that the two placement settings' difference in their spatial distributions of the success rate can be attributed mostly to the difference in the regrasp possibility between the two settings rather than the differences in the robotic arm's kinematic reachability. To show this, in Figure~\ref{fig:spatialsuccessMore}(c) we demonstrate the robotic arm's kinematic reachability, which is computed for a plane at the height of $h$ above the plane where the regrasp experiments are performed. Here, $h$ is the average height for the center of mass of all objects in different placements. The reachability is the same for different objects, but we can observe that the spatial distribution of the success rate is very different between $l$ and $1t1p$ using the planar placement. Such differences are mainly caused by the different regrasp possibilities of these two objects. Similarly, the reachability is roughly the same while using either the planar placement or the pin placement, but the spatial distribution over the success rate is greatly improved for the object $1t1p$ when using the pin placement instead of the planar placement. Such an improvement is mainly due to the improved regrasp possibility of the pin placement compared to that of the planar placement. 
}


\begin{figure}[!t]
\centering
\subfloat[$1t1p$]{\includegraphics[width=\linewidth]{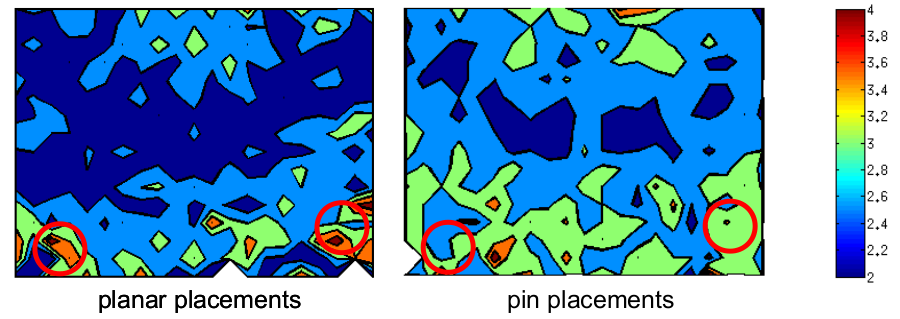}} \\
\subfloat[$l$]{\includegraphics[width=\linewidth]{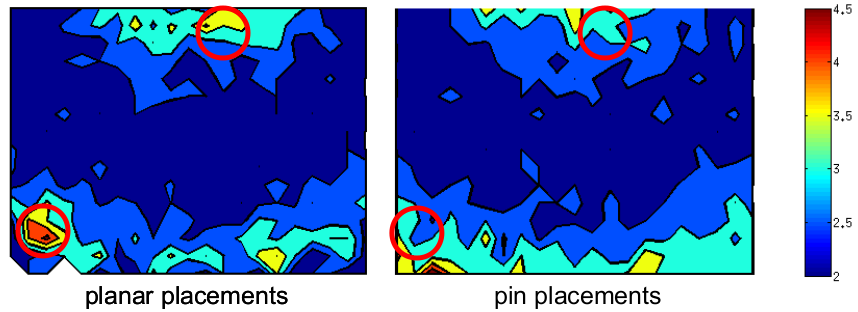}} 
\caption{Comparison of the spatial distribution of the regrasp sequence length
between the pin placement and the planar placement while orienting $1t1p$ and
$l$ models. The robot is standing in the north side of the rectangle working
space. The regions marked by the red circles are challenging workspace where the
success rate is relatively lower for both placements.}
\label{fig:spatiallength}
\end{figure}

Next, we compare two different placement settings' spatial distribution over the
length of the regrasp sequence, as shown in Figure~\ref{fig:spatiallength}. We
can observe in Figure~\ref{fig:spatialsuccess} that for challenging regions where the success rate is relatively
lower for both placements, the sequence
length is shorter for the pin placement compared to the sequence length for the planar placement. This is
because in these situations, reorientation is difficult and the pin placement's highly connected regrasp graph 
is more likely to generate
efficient regrasps. For other places in the center of the workspace where the
success rate is closer to 100\%, both placements have similar average lengths for
regrasp sequences. The sequence using pin placements may be a bit longer
sometimes, also because the placement set is larger for pin cases than for
planar cases.

\subsection{Assembly Tasks}
\begin{figure*}[!p]
\centering
\subfloat[]{\includegraphics[width=0.15\linewidth]{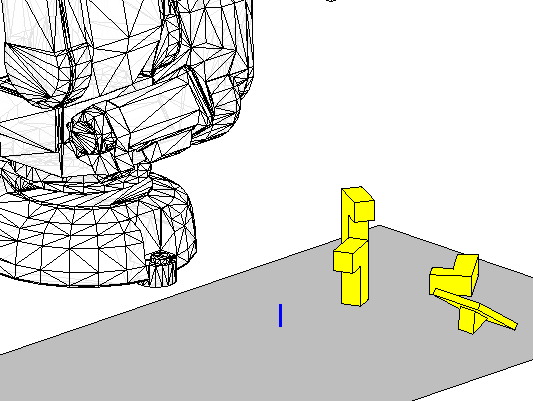}} 
\subfloat[]{\includegraphics[width=0.15\linewidth]{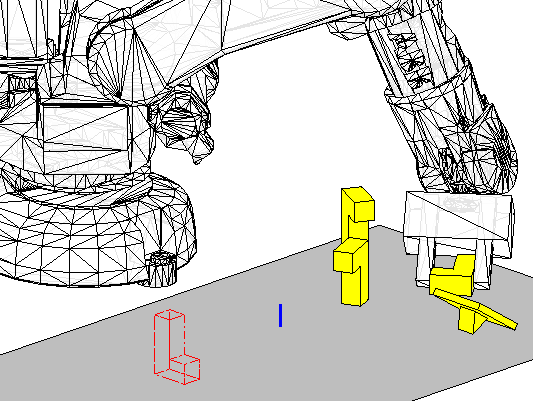}} 
\subfloat[]{\includegraphics[width=0.15\linewidth]{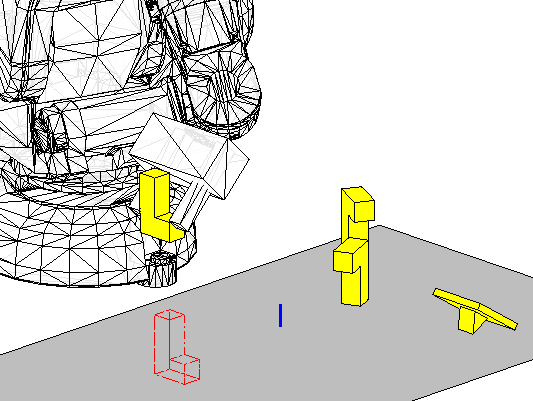}} 
\subfloat[]{\includegraphics[width=0.15\linewidth]{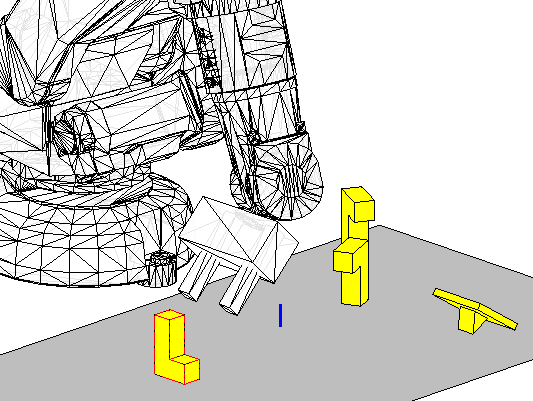}}  
\subfloat[]{\includegraphics[width=0.15\linewidth]{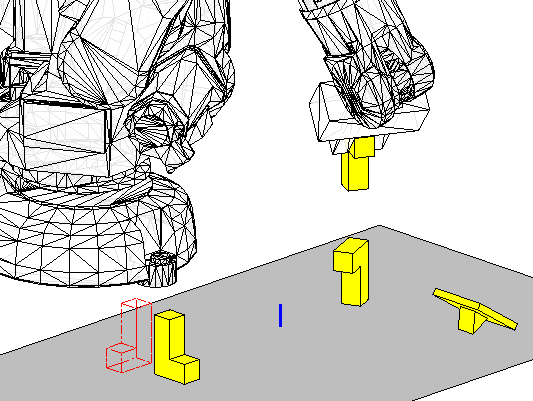}} 
\subfloat[]{\includegraphics[width=0.15\linewidth]{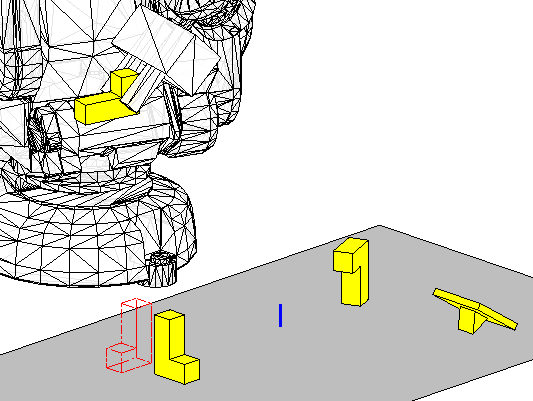}} \\
\subfloat[]{\includegraphics[width=0.15\linewidth]{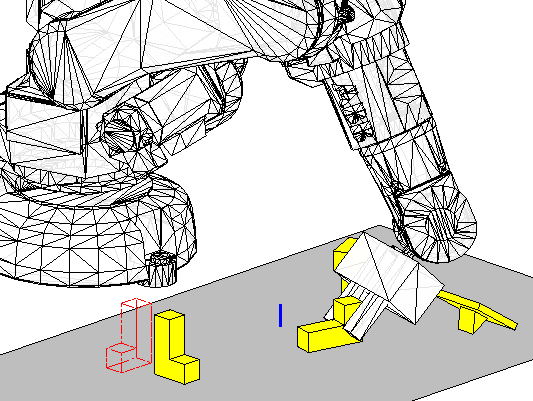}} 
\subfloat[]{\includegraphics[width=0.15\linewidth]{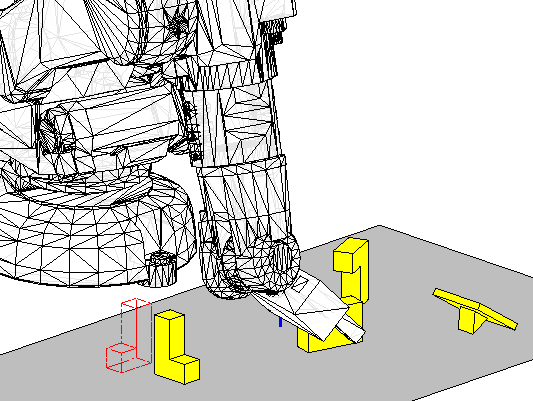}} 
\subfloat[]{\includegraphics[width=0.15\linewidth]{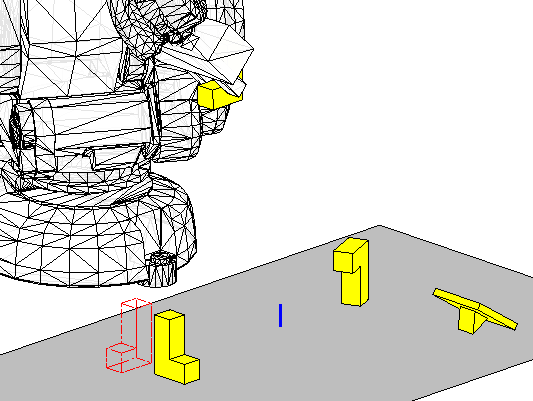}} 
\subfloat[]{\includegraphics[width=0.15\linewidth]{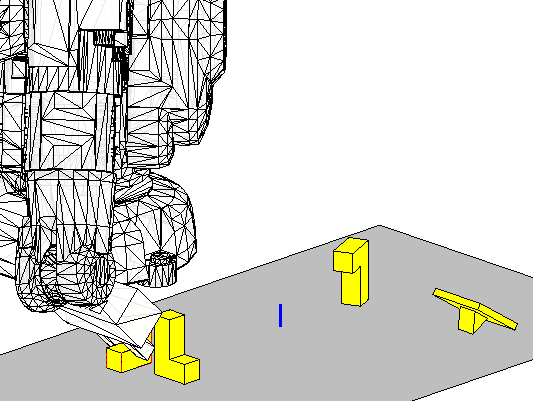}} 
\subfloat[]{\includegraphics[width=0.15\linewidth]{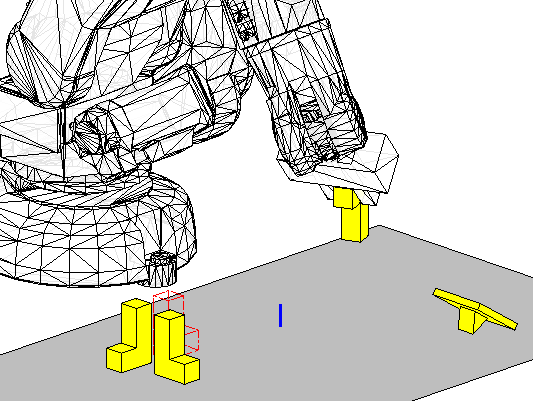}} 
\subfloat[]{\includegraphics[width=0.15\linewidth]{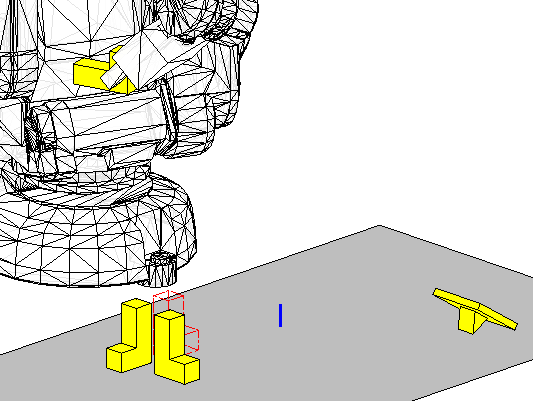}} \\
\subfloat[]{\includegraphics[width=0.15\linewidth]{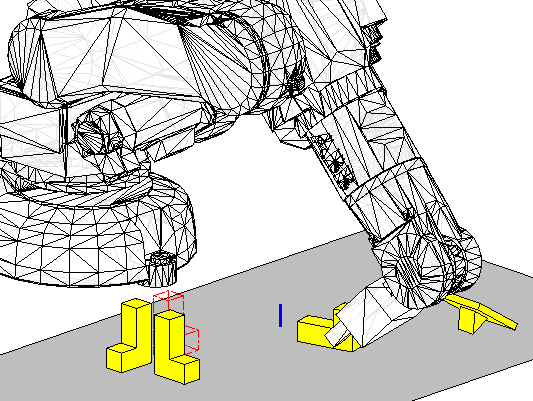}} 
\subfloat[]{\includegraphics[width=0.15\linewidth]{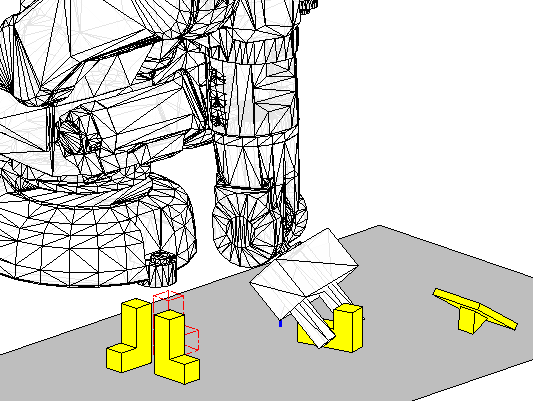}} 
\subfloat[]{\includegraphics[width=0.15\linewidth]{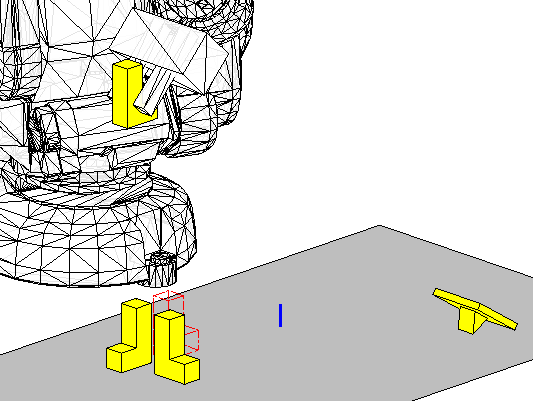}} 
\subfloat[]{\includegraphics[width=0.15\linewidth]{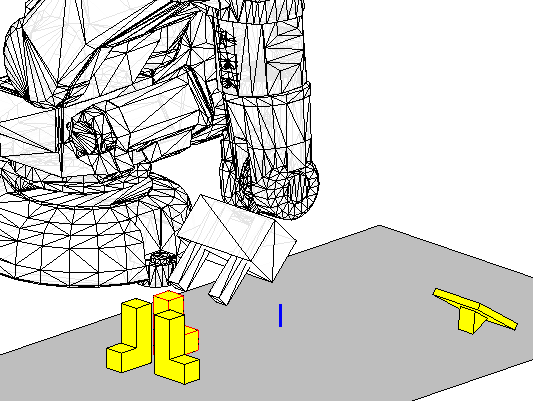}} 
\subfloat[]{\includegraphics[width=0.15\linewidth]{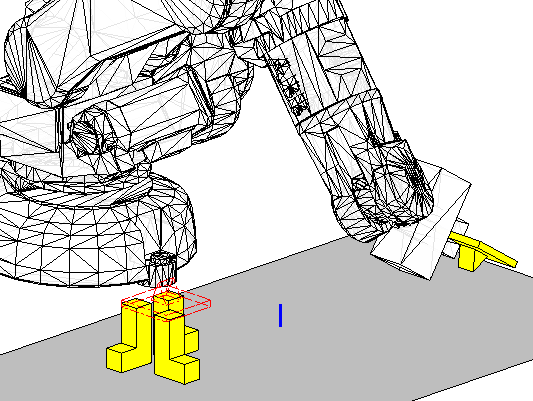}} 
\subfloat[]{\includegraphics[width=0.15\linewidth]{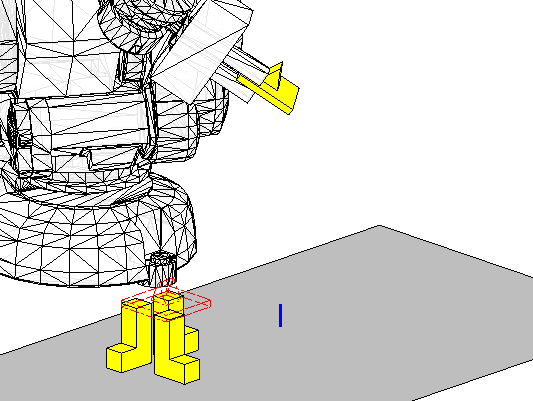}} \\
\subfloat[]{\includegraphics[width=0.15\linewidth]{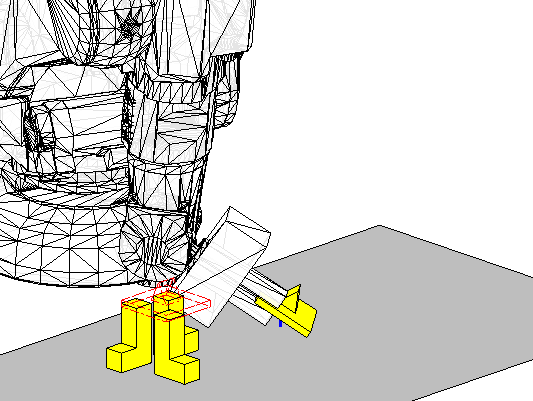}} 
\subfloat[]{\includegraphics[width=0.15\linewidth]{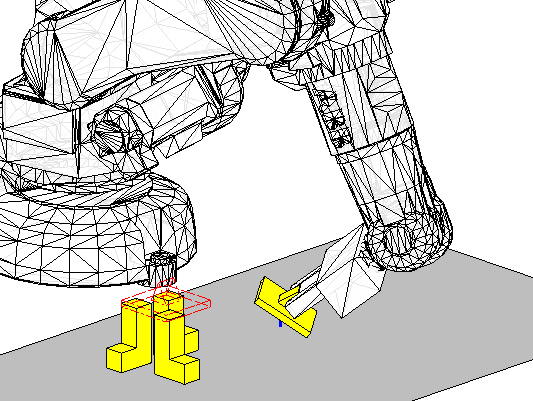}} 
\subfloat[]{\includegraphics[width=0.15\linewidth]{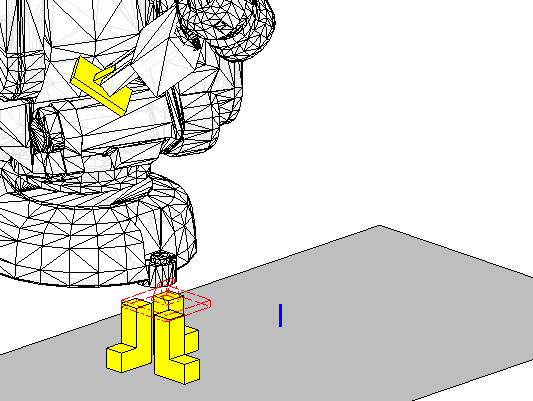}} 
\subfloat[]{\includegraphics[width=0.15\linewidth]{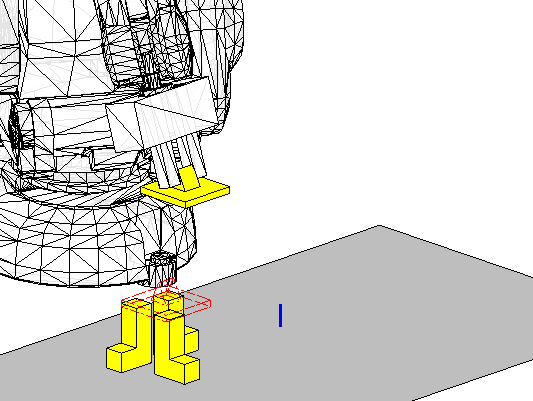}} 
\subfloat[]{\includegraphics[width=0.15\linewidth]{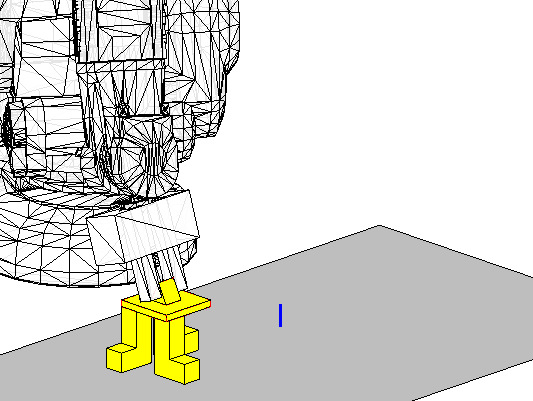}} 
\subfloat[]{\includegraphics[width=0.15\linewidth]{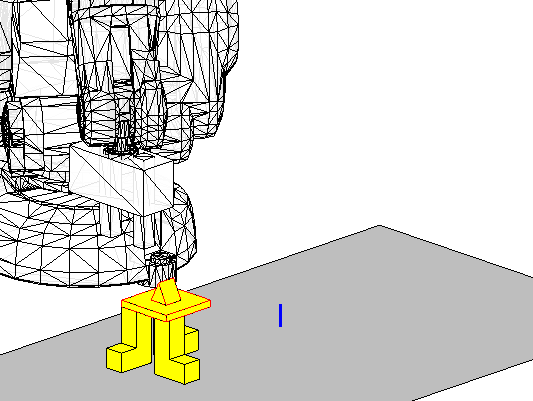}} 
\caption{The regrasp sequence for assembling four different objects, including
one pot lid like object. The sequence is computed with the two-layer regrasp
graph built on the grasps computed by sampling the mesh models. The first object
is oriented without regrasp (b)-(d). All other objects require one step of
regrasp: the second and the third object use the planar placement for
regrasping, and the transition steps occurs at (g)-(h) and (m)-(n),
respectively. The final object is of the pot lid shape, and the robot must
leverage the pin placement to successfully reorient the object, and the
transition happens at (s)-(t).
}
\label{fig:assembly_sequence}
\end{figure*}

Moreover, we perform an assembly task in which the robot picks up
components from one side of the working area and places them down at the other
side to construct a predefined structure. During the process, the robot
chooses suitable re-placement settings for transition on its own, i.e., the
regrasp graph is a mixture of pin placements and planar placements. The
resulting regrasp sequence is shown in Figure~\ref{fig:assembly_sequence}. We
can see that the first object can be directly reoriented without regrasp, while
the next two objects achieve the regrasp only using the planar support. The
final pot lid like object is the most challenging one. The task of placing it on top of the
other components is impossible without using the pin for an intermediate
placement. This task is also achievable by only using the pin placement setting,
but is not possible while only using the planar placement setting. This proves
the efficacy of a support pin placement in challenging assembly tasks.

\section{Conclusion}
In this paper, we improve the pick-and-place regrasp planning by using an
added support pin for intermediate placements. We demonstrate that by using
the pin placements, we can improve the connectivity of the regrasp graph, and
eventually increase the success rate of the reorientation task for objects with
various shapes. We perform a large number of experiments to empirically investigate how the
performance of the reorientation task is influenced by a variety of factors, including
the pin length, the object's size, the object's shape, and the grasp density. 
Moreover, we also show one challenging assembly task which is not possible
without the pin placements.

There are many directions for future work. First, we would like to learn 
a closed form expression for the relation between the reorientation success rate, the pin length, and the object size, by
collecting a large amount of training data from repeated experiments. 
The learned formulation can be used to compute the optimal pin length for a given reorientation task. 
Second, our current regrasp planning framework does not take into the uncertainty that is ubiquitous in the real-world applications.
For instance, our method assumes the validity of the placements and grasps to be solely determined by the geometric shape of the object,
which is not true in reality since the (unknown) physical properties of the object (e.g., whether the object is made of wood, plastic, or light metal like aluminum)
are also important for the grasp quality. We plan to design some robustness criteria for the placements and grasps (e.g., the torque constraints) in order to filter out all unstable grasps and achieve a robust regrasp sequence.  
Eventually, we are interested in implementing the entire framework on real industrial robots for challenging manufacturing tasks such as 3C assembly. 

\bibliographystyle{IEEEtran}
\bibliography{references}

\begin{thebibliography}{10}
\providecommand{\url}[1]{#1}
\csname url@rmstyle\endcsname
\providecommand{\newblock}{\relax}
\providecommand{\bibinfo}[2]{#2}
\providecommand\BIBentrySTDinterwordspacing{\spaceskip=0pt\relax}
\providecommand\BIBentryALTinterwordstretchfactor{4}
\providecommand\BIBentryALTinterwordspacing{\spaceskip=\fontdimen2\font plus
\BIBentryALTinterwordstretchfactor\fontdimen3\font minus
  \fontdimen4\font\relax}
\providecommand\BIBforeignlanguage[2]{{%
\expandafter\ifx\csname l@#1\endcsname\relax
\typeout{** WARNING: IEEEtran.bst: No hyphenation pattern has been}%
\typeout{** loaded for the language `#1'. Using the pattern for}%
\typeout{** the default language instead.}%
\else
\language=\csname l@#1\endcsname
\fi
#2}}

\bibitem{Tournassoud:1987:R}
P.~Tournassoud, T.~Lozano-Perez, and E.~Mazer, ``Regrasping,'' in \emph{IEEE
  International Conference on Robotics and Automation}, vol.~4, 1987, pp.
  1924--1928.

\bibitem{Wan:ICRA:2015}
W.~Wan, M.~Mason, R.~Fukui, and Y.~Kuniyoshi, ``Improving regrasp algorithms to
  analyze the utility of work surfaces in a workcell,'' in \emph{IEEE
  International Conference on Robotics and Automation}, 2015, pp. 4326--4333.

\bibitem{Lozano-Perez:1992:HRT}
T.~Lozano-P{\'e}rez, J.~L. Jones, P.~A. O'Donnell, and E.~Mazer, \emph{Handey:
  A Robot Task Planner}.\hskip 1em plus 0.5em minus 0.4em\relax Cambridge, MA,
  USA: MIT Press, 1992.

\bibitem{Terasaki:1998:MPI}
H.~Terasaki and T.~Hasegawa, ``Motion planning of intelligent manipulation by a
  parallel two-fingered gripper equipped with a simple rotating mechanism,''
  \emph{IEEE Transactions on Robotics and Automation}, vol.~14, no.~2, pp.
  207--219, Apr 1998.

\bibitem{Wan:CASE:2015}
W.~Wan, E.~Cheung, J.~Pan, and K.~Harada, ``Optimizing the parameters of
  tilting surfaces in robotic workcells,'' in \emph{IEEE International
  Conference on Automation Science and Engineering}, 2015.

\bibitem{Harada:2014:VOP}
K.~Harada, T.~Tsuji, K.~Nagata, N.~Yamanobe, and H.~Onda, ``Validating an
  object placement planner for robotic pick-and-place tasks,'' \emph{Robotics
  and Autonomous Systems}, vol.~62, no.~10, pp. 1463--1477, Oct. 2014.

\bibitem{Schuster:2010:Humanoids}
M.~Schuster, J.~Okerman, H.~Nguyen, J.~Rehg, and C.~Kemp, ``Perceiving clutter
  and surfaces for object placement in indoor environments,'' in \emph{IEEE-RAS
  International Conference on Humanoid Robots}, 2010, pp. 152--159.

\bibitem{Lozano:2014:IROS}
T.~Lozano-Perez and L.~Kaelbling, ``A constraint-based method for solving
  sequential manipulation planning problems,'' in \emph{IEEE/RSJ International
  Conference on Intelligent Robots and Systems}, 2014, pp. 3684--3691.

\bibitem{Jiang:2012:LPN}
Y.~Jiang, M.~Lim, C.~Zheng, and A.~Saxena, ``Learning to place new objects in a
  scene,'' \emph{International Journal of Robotics Research}, vol.~31, no.~9,
  pp. 1021--1043, Aug. 2012.

\bibitem{Baumgartl:2014:ICRA}
J.~Baumgartl, T.~Werner, P.~Kaminsky, and D.~Henrich, ``A fast, gpu-based
  geometrical placement planner for unknown sensor-modelled objects and
  placement areas,'' in \emph{IEEE International Conference on Robotics and
  Automation}, 2014, pp. 1552--1559.

\bibitem{Fu:2008:UOM}
H.~Fu, D.~Cohen-Or, G.~Dror, and A.~Sheffer, ``Upright orientation of man-made
  objects,'' \emph{ACM Transactions on Graphics}, vol.~27, no.~3, pp.
  42:1--42:7, 2008.

\bibitem{Kriegman:1997:LTF}
D.~J. Kriegman, ``Let them fall where they may: Capture regions of curved
  objects and polyhedra,'' \emph{International Journal of Robotics Research},
  vol.~16, no.~4, pp. 448--472, 1997.

\bibitem{Moll:2002:MPD}
M.~Moll and M.~A. Erdmann, ``Manipulation of pose distributions,''
  \emph{International Journal of Robotics Research}, vol.~21, no.~3, pp.
  277--292, 2002.

\bibitem{Simeon:2004:IJRR}
T.~Sim\'{e}on, J.-P. Laumond, J.~Cort\'{e}s, and A.~Sahbani, ``Manipulation
  planning with probabilistic roadmaps,'' \emph{The International Journal of
  Robotics Research}, vol.~23, no. 7-8, pp. 729--746, 2004.

\bibitem{Hauser:2011:IJRR}
K.~Hauser and V.~Ng-Thow-Hing, ``Randomized multi-modal motion planning for a
  humanoid robot manipulation task,'' \emph{The International Journal of
  Robotics Research}, vol.~30, no.~6, pp. 678--698, 2011.

\bibitem{Lagriffoul:2012:IROS}
F.~Lagriffoul, D.~Dimitrov, A.~Saffiotti, and L.~Karlsson, ``Constraint
  propagation on interval bounds for dealing with geometric backtracking,'' in
  \emph{IEEE/RSJ International Conference on Intelligent Robots and Systems},
  2012, pp. 957--964.

\bibitem{Dogar:2015:ICRA}
M.~Dogar, A.~Spielberg, S.~Baker, and D.~Rus, ``Multi-robot grasp planning for
  sequential assembly operations,'' in \emph{IEEE International Conference on
  Robotics and Automation}, 2015, pp. 193--200.

\bibitem{Carlisle:1994:PGF}
B.~Carlisle, K.~Goldberg, A.~Rao, and J.~Wiegley, ``A pivoting gripper for
  feeding industrial parts,'' in \emph{IEEE International Conference on
  Robotics and Automation}, 1994, pp. 1650--1655.

\bibitem{Yoshida:2010:PBM}
E.~Yoshida, M.~Poirier, J.-P. Laumond, O.~Kanoun, F.~Lamiraux, R.~Alami, and
  K.~Yokoi, ``Pivoting based manipulation by a humanoid robot,'' \emph{Auton.
  Robots}, vol.~28, no.~1, pp. 77--88, Jan. 2010.

\bibitem{Lynch:1996:SPM}
K.~Lynch and M.~Mason, ``Stable pushing: Mechanics, controllability, and
  planning,'' \emph{International Journal of Robotics Research}, 1996.

\bibitem{Chavan:2015:PPI}
N.~Chavan-Dafle and A.~Rodriguez, ``Prehensile pushing: In-hand manipulation
  with push-primitives,'' in \emph{IEEE/RSJ International Conference on
  Intelligent Robots and Systems}, 2015.

\bibitem{Chavan:2014:EDI}
N.~Chavan-Dafle, A.~Rodriguez, R.~Paolini, B.~Tang, S.~S. Srinivasa,
  M.~Erdmann, M.~T. Mason, I.~Lundberg, H.~Staab, and T.~Fuhlbrigge,
  ``Extrinsic dexterity: In-hand manipulation with external forces,'' in
  \emph{IEEE International Conference on Robotics and Automation}, 2014.

\bibitem{Lee:2015:HPM}
G.~Lee, T.~Lozano-Perez, and L.~P. Kaelbling, ``Hierarchical planning for
  multi-contact non-prehensile manipulation,'' in \emph{IEEE/RAS International
  Conference on Intelligent Robots and Sytems}, 2015.

\bibitem{Jentzsch:2015:MOPL}
S.~Jentzsch, A.~Gaschler, O.~Khatib, and A.~Knoll, ``{MOPL}: A multi-modal path
  planner for generic manipulation tasks,'' in \emph{IEEE/RSJ International
  Conference on Intelligent Robots and Systems}, 2015.

\bibitem{Koga:1994:OMA}
Y.~Koga and J.-C. Latombe, ``On multi-arm manipulation planning,'' in
  \emph{IEEE International Conference on Robotics and Automation}, 1994, pp.
  945--952.

\bibitem{Vahrenkamp:2009:HMP}
N.~Vahrenkamp, D.~Berenson, T.~Asfour, J.~Kuffner, and R.~Dillmann, ``Humanoid
  motion planning for dual-arm manipulation and re-grasping tasks,'' in
  \emph{IEEE/RSJ International Conference on Intelligent Robots and Systems},
  2009, pp. 2464--2470.

\bibitem{Cohen:2012:SPD}
B.~Cohen, S.~Chitta, and M.~Likhachev, ``Search-based planning for dual-arm
  manipulation with upright orientation constraints,'' in \emph{IEEE
  International Conference on Robotics and Automation}, 2012, pp. 3784--3790.

\bibitem{Dobson:2015:PRA}
A.~Dobson and K.~E. Bekris, ``Planning representations and algorithms for
  prehensile multi-arm manipulation,'' in \emph{IEEE/RAS International
  Conference on Intelligent Robots and Sytems}, 2015.

\bibitem{Cohen:2015:PSM}
B.~Cohen, M.~Phillips, and M.~Likhachev, ``Planning single-arm manipulations
  with n-arm robots,'' in \emph{Robotics: Science and System}, 2015.

\end{thebibliography}

\end{document}